\documentclass[sigconf]{acmart}
\usepackage[utf8]{inputenc}
\usepackage{newunicodechar}
\usepackage{textcomp}
\usepackage[utf8]{inputenc}
\AtBeginDocument{%
  }

\setcopyright{acmlicensed}
\copyrightyear{2018}
\acmYear{2018}
\acmDOI{XXXXXXX.XXXXXXX}
\acmConference[MM'26]{}{November 10--14,
  2026}{Rio de Janeiro, Brazil}
\renewcommand\footnotetextcopyrightpermission[1]{}
\renewcommand\copyrightyear{}
\renewcommand\acmYear{}
\renewcommand\acmConference[2]{}
\renewcommand\acmBooktitle{}
\renewcommand\acmPrice{}
\renewcommand\acmDOI{}
\renewcommand\acmISBN{}

\usepackage{amsmath}
\usepackage{algorithm}
\usepackage{algpseudocode}
\usepackage{multirow}
\usepackage{booktabs}
\usepackage{graphicx}
\usepackage[table]{xcolor} 
\usepackage{tabularx}
\usepackage{array}
\usepackage{geometry}
\usepackage{amsfonts} 
\definecolor{tablegray}{HTML}{EFEFEF} 
\algtext*{EndFor}
\algtext*{EndIf}
\begin{document}

\title[ReliableRAG]{ReliableRAG: Combating Misinformation in Retrieval-Augmented Generation via Reliability-Guided Reasoning Chains}

\author{Jinpu Jiang}
\authornote{Equal Contribution.}
\affiliation{%
  \institution{College of Software, Jilin University}
  \city{Changchun}
  \state{Jilin}
  \country{China}
}
\email{jiangjp24@mails.jlu.edu.cn}

\author{Xuan Wu}
\authornotemark[1]
\affiliation{%
  \institution{College of Computer Science and Technology, Jilin University}
  \city{Changchun}
  \state{Jilin}
  \country{China}
}
\email{wuuu22@mails.jlu.edu.cn}

\author{Wenhao Song}
\affiliation{%
  \institution{College of Software, Jilin University}
  \city{Changchun}
  \state{Jilin}
  \country{China}
}
\email{songwh23@mails.jlu.edu.cn}

\author{Bo Yang}
\affiliation{%
  \institution{Key Laboratory of Symbolic Computation and Knowledge Engineering of Ministry of Education, College of Computer Science and Technology, Jilin University}
  \city{Changchun}
  \state{Jilin}
  \country{China}
}
\email{ybo@jlu.edu.cn}

\author{You Zhou}
\authornotemark[2]
\affiliation{%
  \institution{Key Laboratory of Symbolic Computation and Knowledge Engineering of Ministry of Education, College of Computer Science and Technology, Jilin University}
  \city{Changchun}
  \state{Jilin}
  \country{China}
}
\email{zyou@jlu.edu.cn}

\author{Hongwei Ge}
\affiliation{%
  \institution{College of Computer Science and Technology, Dalian University of Technology}
  \city{Dalian}
  \state{Liaoning}
  \country{China}
}
\email{hwge@dlut.edu.cn}

\author{Heow Pueh Lee}
\affiliation{%
  \institution{Department of Mechanical Engineering, National University of Singapore}
  \country{Singapore}
}
\email{mpeleehp@nus.edu.sg}

\author{Yanchun Liang}
\affiliation{%
  \institution{School of Computer Science, Zhuhai College of Science and Technology}
  \city{Zhuhai}
  \state{Guangdong}
  \country{China}
}
\email{ycliang@jlu.edu.cn}

\author{Chunguo Wu}
\authornote{Corresponding Authors.}
\affiliation{%
  \institution{Key Laboratory of Symbolic Computation and Knowledge Engineering of Ministry of Education, College of Computer Science and Technology, Jilin University}
  \city{Changchun}
  \state{Jilin}
  \country{China}
}
\email{wucg@jlu.edu.cn}

\renewcommand{\shortauthors}{Jiang et al.}

\begin{abstract}
Retrieval-Augmented Generation (RAG) has emerged as a potent system architecture for addressing Question Answering (QA) tasks by integrating external information into Large Language Models (LLMs). However, the proliferation of false, inaccurate, and misleading information in news and social media poses a challenge to real-world RAG systems, particularly in multi-hop QA where complex, multi-step reasoning is easily misled by even a single deceptive misinformation segment within the misinformation-polluted source documents. While existing approaches primarily adopt paradigms of implicit alignment or explicit regulation to mitigate this issue, their inability to perceive fine-grained information reliability makes them highly susceptible to fine-grained deceptive misinformation that is semantically aligned with the question but factually incorrect, ultimately leading to the generation of erroneous answers. To address this limitation, we propose ReliableRAG, which, to the best of our knowledge, is the first reliability-driven framework designed to prevent the misleading effects of deceptive misinformation in multi-hop QA through the evaluation of fine-grained individual triples. Specifically, ReliableRAG first extracts information segments from source documents and represents them as fine-grained structured triples. It then performs fine-grained reliability quantification by synthesizing query-triple semantic relevance and triple credibility, ensuring the retention of only the top-$K$ most reliable and non-redundant triples. By autoregressively constructing robust reasoning chains from these refined triples, the framework effectively consolidates trustworthy information and filters out deceptive misinformation, ensuring accurate answers that is faithful to reliable information. Experimental results on three multi-hop QA datasets demonstrate that ReliableRAG outperforms existing methods, significantly enhancing the factual reliability and robustness of RAG systems under deceptive misinformation injection.
\end{abstract}



\maketitle

\section{Introduction}
Retrieval-Augmented Generation (RAG), a potent system architecture for Question Answering (QA) tasks, empowers Large Language Models (LLMs) by incorporating external information \cite{RF4,RF111,RF28}. To resolve multi-hop QA tasks that require complex, multi-step reasoning \cite{RF24,RF25,RF26}, this system requires retrieving the top-$K$ documents most relevant to the given question $Q$ from a vast Web corpus to aggregate complementary information across disparate sources. These documents are then prepended to the question, with the resulting concatenation serving as the input to an LLM-based generator to produce the final answer \cite{RF46}. However, the reliability of RAG can be substantially compromised when the retrieved documents contain maliciously generated deceptive misinformation, ultimately degrading their overall performance \cite{RF9,RF10,RF105,RF106}, particularly in multi-hop Question Answering (QA) tasks. In such scenarios, even a few deceptive misinformation segments may propagate through the reasoning process, triggering a cascade of errors that corrupts the entire reasoning chain and yields incorrect answers \cite{RF81,RF64,RF37,RF63}.

\begin{figure}[!t]
\centerline{\includegraphics[width=\columnwidth]{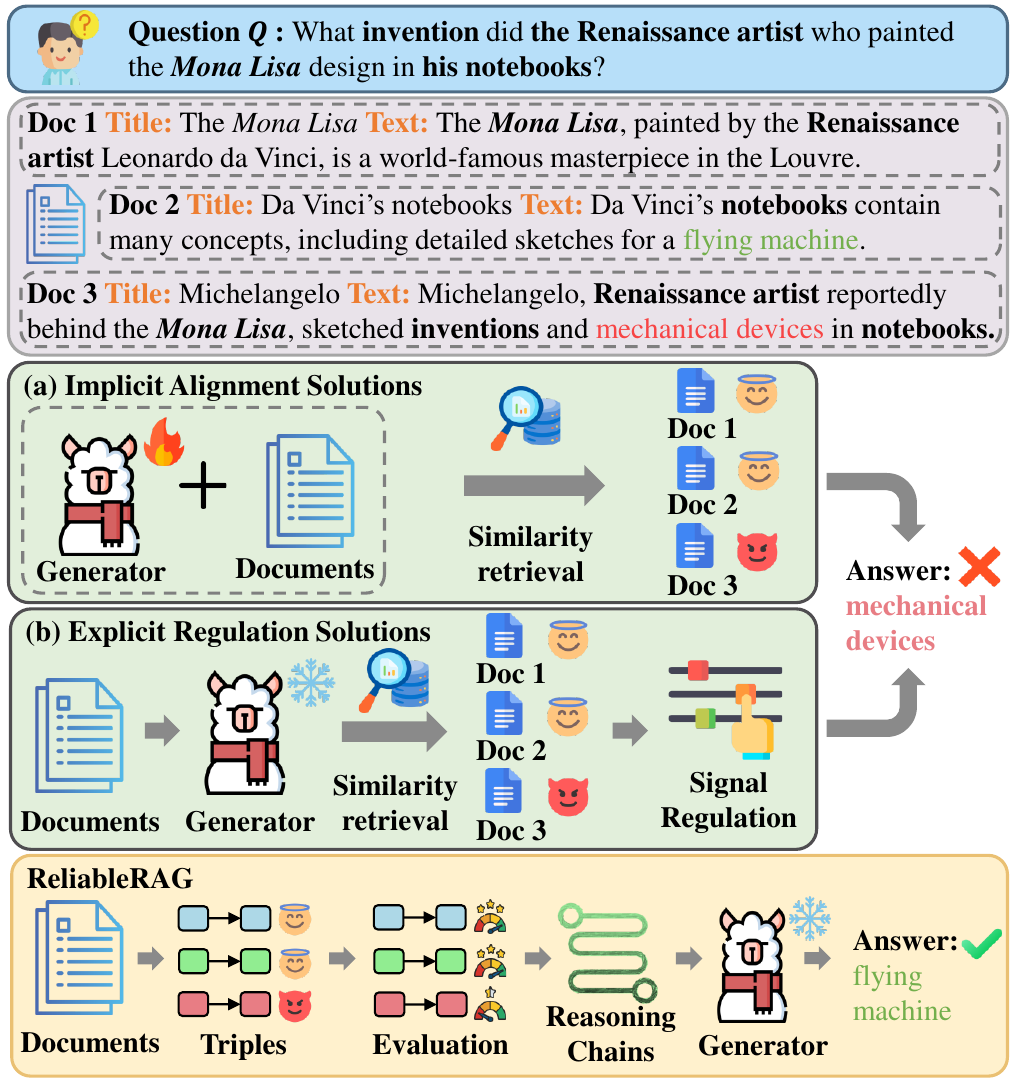}}
\caption{Illustration of the differences between existing methods and ReliableRAG under the interference of misinformation. Green text denotes the correct answer, while red text represents a misleading incorrect answer. While implicit alignment and explicit regulation paradigms often falter when faced with semantically similar fine-grained deceptive misinformation, thereby leading to incorrect answers, ReliableRAG identifies the fine-grained reliability of information and constructs robust reasoning chains to prevent deceptive misinformation, ensuring accurate answers.}
\label{fig:1}
\end{figure} 

To improve the robustness of RAG facing misinformation in multi-hop QA, Recent studies primarily adopt two  paradigms, namely implicit alignment and explicit regulation. Specifically, implicit alignment methods achieve this by training models to internalize criteria such as self-reflection signals, learned decision policies, or credibility awareness, enabling them to autonomously regulate reliance on retrieved information during generation and thereby enhance factual accuracy in retrieval-augmented systems \cite{RF68, RF11, RF67}.  However, these methods necessitate substantial computational resources and high-quality training data which are often difficult to acquire \cite{RF118}. Moreover, the inherent domain gap between specialized training datasets and real-world test scenarios limits their generalization \cite{RF117}. In contrast, explicit regulation approaches explicitly regulate the model's reliance on retrieved information using external signals such as document credibility scores (i.e., the extent to which a document is free from misinformation) or knowledge graph structures, thereby enhancing factual accuracy and robustness in RAG systems \cite{RF22, RF66}. Despite these advancements, current explicit regulation methods exhibit critical shortcomings in perceiving fine-grained information reliability. As illustrated in Figure~\ref{fig:1}, these approaches often fail to discern between highly relevant but factually incorrect fine-grained deceptive misinformation, leading them to be misled by such misinformation. These misinformation subsequently propagate through and undermine the entire reasoning chain, ultimately yielding inaccurate answers \cite{RF81, RF65}. Therefore, in this work, we investigate the following research question:

\noindent \textit{Can we equip RAG systems with fine-grained reliability awareness to discern deceptive misinformation within fine-grained information, and thereby prevent its propagation during multi-hop reasoning?}
 
To address this research question, we propose ReliableRAG, which, to the best of our knowledge, is the first reliability-driven framework designed to prevent the misleading effects of deceptive misinformation in multi-hop QA through the evaluation of fine-grained individual triples. 
The framework achieves this through three integrated modules: Triple Extraction, Triple Evaluation, and Chain Construction, which operate across two distinct phases. In the offline phase, the Triple Extraction module pre-processes each source document into fine-grained structured triple representations. Subsequently, an LLM-based evaluator is employed to pre-assess these triples and assign each an interpretable credibility score (i.e., the extent to which a triple is free from misinformation). Collectively, these steps provide fine-grained information and pre-computed factors for subsequent retrieval and inference. During online phase, given the question $Q$, the Triple Evaluation module first dynamically formulates a unique search query for every individual existing reasoning chain constructed from the preceding hop by prepending its accumulated information to the question $Q$. This one-to-one mapping ensures that each query is uniquely tailored to its corresponding chain to retrieve the supplementary information required to extend that chain further. Leveraging these queries, the module quantifies the reliability of each triple by synthesizing two factors: query-triple semantic relevance and triple credibility. Based on these fine-grained reliability scores, the module selects the top-$K$ most reliable and non-redundant triples for each existing chain, thereby filtering out those containing deceptive misinformation to provide a highly trustworthy search space for constructing candidate reasoning chains for the current hop. Operating within this search space, the Chain Construction module employs a beam search strategy to autoregressively construct multiple candidate reasoning chains. Specifically, the module employs a multiple-option reasoning prompt to guide an LLM-based selector in deciding, for each existing chain, whether to append the most probable triple from the search space to expand it, or to execute an early termination if the accumulated information within it is already sufficient to answer $Q$. Simultaneously, the module updates the confidence score of each candidate chain based on the probability of the selector's decision. According to these updated scores, only the top-$F$ most confident candidate chains are retained as the input for the subsequent hop. In this manner, the iterative cycle of Triple Evaluation and Chain Construction continues until all chains terminate or a maximum hop limit is reached. Ultimately, the framework synthesizes the final robust chains into a refined and reliable supporting context, denoted as $\mathcal{Z}$. This $\mathcal{Z}$ is prepended to the question $Q$ to form a unified input $\mathcal{X}$, which is fed into the generator to produce the final answer.

To evaluate the effectiveness of ReliableRAG, we conduct extensive experiments on three multi-hop QA datasets adversarially augmented with deceptive misinformation. The experimental results demonstrate that ReliableRAG outperforms existing methods across all datasets, effectively mitigating the interference of misinformation. Further analyses indicate that by anchoring reasoning in reliable triples, our framework enhances the factual reliability and robustness of the RAG system, enabling more accurate answers even under deceptive misinformation injection.
\par
The key contributions of this work are as follows.
\par
\textbf{I)} We discover that prioritizing fine-grained reliability at the triple level facilitates the filtering of fine-grained deceptive misinformation, thereby preventing its propagation and promoting more robust and reliable multi-hop reasoning.

\textbf{II)} We propose ReliableRAG, which, to the best of our knowledge, is the first reliability-driven framework that features a novel Triple Evaluation module to dynamically quantify fine-grained information reliability at the triple level, thus preventing the misleading effects of deceptive misinformation in multi-hop QA and constructing a highly trustworthy search space.

\textbf{III)} ReliableRAG seamlessly integrates the Triple Evaluation module into the reasoning pipeline, enabling the Chain Construction module to leverage a highly trustworthy search space to autoregressively construct robust, multi-hop reasoning chains that are anchored in reliable triples, thereby preventing misinformation propagation before generating the final answer.

\textbf{IV)} We conduct extensive experiments on three multi-hop QA datasets adversarially augmented with deceptive misinformation. The results demonstrate that ReliableRAG significantly outperforms existing methods, thereby enhancing the factual reliability and robustness of RAG systems and validating its effectiveness in systematically preventing misinformation propagation throughout the reasoning process.



\section{Related Work}
RAG advancements in multi-hop QA improve performance by facilitating the reasoning process through techniques such as iterative search, Reasoning with Attributions, and preference optimization \cite{RF42,RF84,RF85}. However, these systems are vulnerable to misinformation, where plausible distractors or adversarial passages can significantly degrade performance and poison the generation process \cite{RF13,RF81,RF86,RF87}. To address this, prior work has evolved from early feature-based classification methods \cite{RF15,RF16,RF47} to modern LLM-based credibility assessment approaches \cite{RF17,RF18} to detect misinformation. Based on this, recent studies have advanced robustness through two primary paradigms: implicit alignment and explicit regulation \cite{RF66,RF22,RF67,RF68,RF11}.

\begin{figure*}[!t]
\centerline{\includegraphics[width=\textwidth]{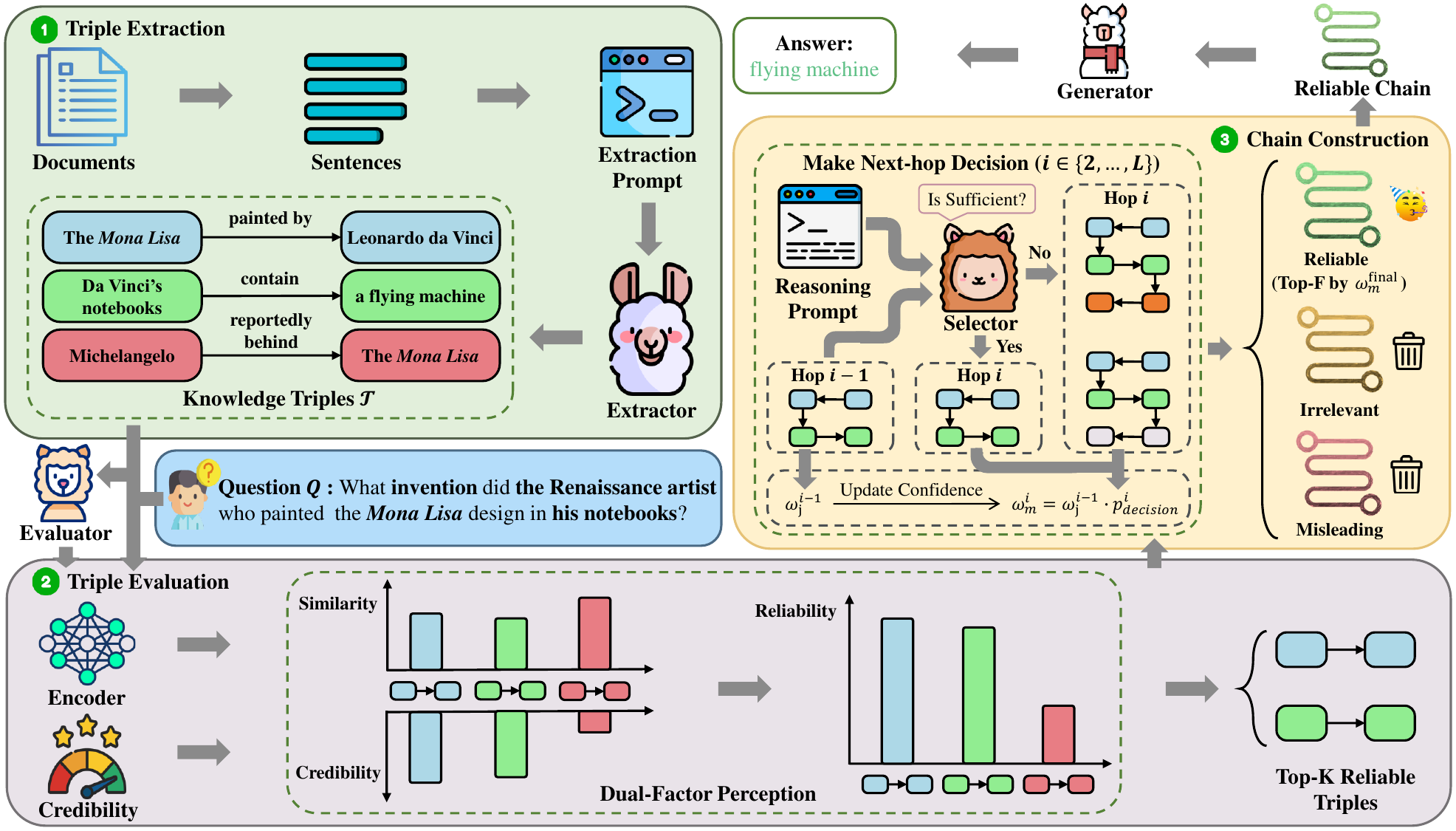}}
\caption{The overall pipeline of the ReliableRAG framework. ReliableRAG first employs the Triple Extraction module to extract information segments from documents and represent them as structured triples. Subsequently, the Triple Evaluation module performs fine-grained reliability quantification by synthesizing query-triple semantic relevance and triple credibility, ensuring the retention of only the top-$K$ most reliable and non-redundant triples for robust multi-hop reasoning. Leveraging these refined triples, the Chain Construction module autoregressively constructs robust reasoning chains to consolidate trustworthy information and filter out deceptive misinformation, ultimately ensuring the generation of accurate answers. 
}
\label{fig:2}
\end{figure*}

Implicit alignment methods achieve this by training models to internalize criteria such as self-reflection signals, learned decision policies, or credibility awareness, enabling them to autonomously regulate reliance on retrieved information during generation and thereby enhance factual accuracy in retrieval-augmented systems. For instance, Asai et al. \cite{RF68} introduced a framework that generates reflection tokens to adaptively judge retrieval necessity and output relevance, thereby filtering out erroneous content. Pan et al. \cite{RF11} employed instruction fine-tuning to teach LLMs to differentiate and process information according to provided credibility indicators. Similarly, Lin et al. \cite{RF67} utilized a reinforcement learning approach to balance internal parametric knowledge with external context, allowing the LLM to fall back on its own knowledge when encountering misleading information. However, these methods necessitate substantial computational resources and high-quality training data which are often difficult to acquire \cite{RF118}. Moreover, the inherent domain gap between specialized training datasets and real-world test scenarios limits their generalization \cite{RF117}.
In contrast, explicit regulation approaches explicitly regulate the model's reliance on retrieved information using external signals such as credibility scores or knowledge graph structures, thereby enhancing factual accuracy and robustness in RAG systems. Deng et al. \cite{RF22} mitigated the influence of low-credibility documents by attenuating their corresponding attention weights during the decoding process. However, this approach relies on coarse-grained, document-level credibility scores, failing to perceive the fine-grained information reliability required for robust multi-hop QA. Furthermore, Liu et al. \cite{RF66} proposed constructing knowledge graphs to identify factual-level conflicts via entropy-based filtering, effectively blocking the propagation of misinformation. Nevertheless, this method retrieves fine-grained structured triples based solely on semantic similarity, which may fail to discern fine-grained deceptive misinformation that is semantically aligned with the question but factually incorrect. Such misinformation subsequently propagates through and undermines the entire reasoning chain, ultimately yielding inaccurate answers \cite{RF81, RF65}.


To address these limitations, we introduce ReliableRAG, a training-free framework that transforms documents into structured triples to quantify fine-grained reliability via query-triple semantic relevance and triple credibility, constructing robust reasoning chains to consolidate trustworthy information and filter out deceptive misinformation.
\section{\textbf{ReliableRAG}}
\label{ReliableRAG}
Given a multi-hop question $Q$, the objective is to derive a correct answer $a$ by identifying and integrating the top-$K$ documents most relevant to $Q$ from the given set of documents $\mathcal{D} = \{d_1, d_2, \dots, d_N\}$, within at most $L$ reasoning hops. These documents are pre-retrieved from external sources such as Wikipedia \cite{RF42}, where each document $d_j$ comprises a title and the corresponding text. However, RAG is vulnerable to misinformative documents that are semantically relevant to $Q$ but factually incorrect, which can mislead the LLM-based generator $\mathcal{M}_{\text{gen}}$ into producing erroneous answers. To address this limitation, as illustrated in Figure~\ref{fig:2}, we introduce ReliableRAG, which comprises three core modules, namely Triple Extraction, Triple Evaluation, and Chain Construction. The source code of ReliableRAG is available online anonymously\footnote{https://anonymous.4open.science/r/ReliableRAG-734C}.

\subsection{Triple Extraction}
 

 
Although LLMs are capable of processing long contexts, directly reasoning over coarse-grained documents often suffers from the lost-in-the-middle phenomenon \cite{RF109}, wherein LLMs tend to overlook critical information located in the middle of long inputs. This phenomenon is further exacerbated in multi-hop QA tasks, where relevant evidence is scattered across multiple documents, thereby making it easier for the generator to overlook key information and hindering the production of accurate answers.

To address this limitation, we incorporate this module that transforms each document $d_i \in \mathcal{D}$, consisting of a title and text content, into multiple fine-grained structured triples, each comprising a subject, predicate, and object. Specifically, we leverage the In-Context Learning (ICL) ability of LLMs \cite{RF94, RF53} to prompt the LLM-based extractor $\mathcal{M}_{\text{ext}}$ to extract triples (see Appendix~\ref{Im_ext} for experiments on various extractors). To facilitate fine-grained processing, the text content of target document $d_i$ is first segmented into individual sentences based on periods. Following recent studies on knowledge graph construction \cite{RF29, RF107, RF108}, the extraction process is guided by an extraction prompt that incorporates both the document $d_i$ (its title and segmented sentences) and three in-context exemplars. These three exemplars are the ones most semantically similar to the title and text of $d_i$, selected from a manually annotated set while ensuring no overlap with the test datasets. Given this prompt, the extractor $\mathcal{M}_{\text{ext}}$ identifies the title, along with information segments within each sentence, as entities, subsequently inferring the relationships between them. This strategy leverages the inherent relevance between the title and each sentence within the text content, ensuring that both the extracted entities and their inferred relationships are substantively anchored to the core topic. Through this parallelized process, a single document yields multiple structured triples, which collectively form the set of triples $\mathcal{T}$ for the question $Q$: 
\begin{align} \mathcal{T} = \{t_j = (s_j, p_j, o_j)\}_{j=1}^M \end{align} 
where $s_j, p_j, o_j$ denote the subject, predicate, and object of the $j$-th triple, respectively, with $M \ge K$. By adopting this design, this module produces a structured set of triples, thereby mitigating the lost-in-the-middle issue. Ablation studies demonstrate that this module significantly contributes to the overall performance (see Section~\ref{exp:abl_sty} for details).

\subsection{Triple Evaluation}
In this subsection, we delineate the process of Triple Evaluation, which comprises Chain Definition and Query Formulation followed by Fine-grained Reliability Quantification.
\subsubsection{Chain Definition and Query Formulation}
To search for the required information to autoregressively construct reasoning  chains, we follow the query formulation in prior studies \cite{RF119,RF42,RF29}. Let $\mathcal{C}^i = \{u^i_j\}^{|\mathcal{C}^i|}_{j=1}$ denote the set of reasoning chains maintained at hop $i$, where $|\mathcal{C}^i|$ represents the current number of reasoning chains. Notably, the superscript $i$ indicates the current reasoning hop rather than the actual depth of each individual chain. During online inference, the process begins at hop $i=1$ with a primary query $q^1$ is first initialized solely from the question $Q$  to trigger the construction of the initial reasoning chains $\mathcal{C}^1$. For subsequent hops $i \in \{2, \dots, L\}$, a set of dynamic queries $\{{q^i_j}\}_{j=1}^{|\mathcal{C}^{i-1}|}$ is generated based on the existing chains in $\mathcal{C}^{i-1}$. Specifically, each query $q^i_j$ is formed by prepending its corresponding reasoning chain $u^{i-1}_j$ to the question $Q$, thereby representing the triple information required at current hop $i$ to further address $Q$.

\subsubsection{Fine-grained Reliability Quantification}
Since the documents contain deceptive misinformation, the extracted triples are also prone to inheriting misleading misinformation. These triples can propagate through the reasoning process, triggering a cascade of errors that corrupts the entire reasoning chain and ultimately leads to incorrect answers \cite{RF81,RF64,RF37,RF63}. To address this, we propose a dual-factor perception mechanism that formulates a fine-grained reliability score by synthesizing the triple's semantic relevance to query $q^i_j$ with triple credibility (i.e., the extent to which the triple is free from misinformation), thereby selecting the top-$K$ most reliable triples from $\mathcal{T}$ for each $q^i_j$ at each hop.
To obtain the fine-grained reliability score, we first measure the semantic relevance of each triple to the query $q^i_j$. 
Specifically, we employ a bi-encoder (see Appendix~\ref{apx:bi_encoder_selection} for experiments on different bi-encoder) to map $q^i_j$ and each triple $t_k \in \mathcal{T}$ into a unified feature space. Each triple $t_k$ is represented as a formatted string $\langle s_k; p_k; o_k \rangle$. By encoding both the query and these formatted strings, we obtain the dense vector representations $h^i_j, h_k \in \mathbb{R}^H$, where $H$ denotes the dimensionality of the embedding space. The semantic relevance is then quantified by the cosine similarity between these vectors, reflecting how closely the information within the triple matches the query's intent: \begin{align}
\Phi(h^i_j, h_k)= \frac {h^i_j \cdot h_k}{\| h^i_j \|  \| h_k \|}
\end{align} A higher value of $\Phi(h^i_j, h_k)$ indicates a stronger semantic match. However, high semantic relevance alone does not guarantee the reliability of selected triples, as semantically relevant triples may still be factually incorrect. This motivates the need for an additional criterion to assess their credibility.
Unlike recent studies that assign credibility scores to entire documents \cite{RF22, RF11}, we propose a fine-grained assessment framework that operates at the triple level. Specifically, this framework evaluates each triple $t_k = (s_k, p_k, o_k)$ based on its basic components: the subject, the predicate, and the object. It employs the LLM-based evaluator $\mathcal{M}_{\text{eva}}$ to perform fine-grained analyses on the credibility of these components (see Appendix~\ref{apx:eva} for experiments on different evaluators). Based on this qualitative analysis, the evaluator $\mathcal{M}_{\text{eva}}$ subsequently derives an interpretable credibility score, denoted as $\eta_k$, for each triple, which is assigned as an additional, respective attribute. This process consists of the following steps:

\textbf{Step 1:} Entity Credibility Analysis. The subject $s_k$ and object $o_k$ are first independently analyzed to determine whether they represent authentic and established entities rather than spurious or hallucinated concepts. This stage provides the analytical basis regarding the factual existence of components for the subsequent evaluation. 

\textbf{Step 2:} Relational Credibility Analysis.  Building upon the entity analysis, the relationship described by the predicate $p_k$ between subject $s_k$ and object $o_k$ is analyzed to verify its authentic coherence. For instance, in the triple \textit{(Albert Einstein, was the first recipient in 1921 of, the Nobel Prize in Physics)}, the analysis recognizes that while the entities are authentic, the specific relational claim is spurious. This provides the logical rationale for determining the triple’s credibility. 

\textbf{Step 3:} Synthesis and Quantitative Scoring. Finally, the qualitative insights from the preceding steps are integrated to generate a finalized credibility score ranging from 0 to 10. This synthesis ensures that the numerical result is grounded in the established analytical steps, reflecting a comprehensive judgment of both entity and relational credibility.

The aforementioned analytical steps are performed by leveraging the ICL ability of the LLM. By employing prompts with few-shot demonstrations, the reasoning behavior of the evaluator $\mathcal{M}_{\text{eva}}$ is constrained to align with this triple-level assessment framework's logic. To enhance scalability, the framework utilizes parallel invocations for large-scale assessment by pre-assigning credibility scores to the triples $\mathcal{T}$ associated with each question $Q$ in an offline manner. The specific prompt structure and design for this assessment are detailed in Appendix~\ref{apx:prompt_stn}. Upon deriving the semantic relevance $\Phi(h^i_j, h_k)$ between each query $q^i_j$ and each triple $t_k \in \mathcal{T}$, alongside the per-triple credibility score $\eta_k$, the dual-factor perception mechanism synthesizes these factors into the fine-grained reliability score. Formally, this fine-grained reliability score is computed as: \begin{align}    
R^i_{j,k} = \alpha \cdot \Phi(h^i_j, h_k) + (1 - \alpha) \cdot \eta_k 
\end{align} where $\alpha \in [0, 1]$ is a balancing coefficient. As each query $q^i_j$ evolve at each hop $i$, the semantic relevance $\Phi(h^i_j, h_k)$ is updated and the fine-grained reliability score $R^i_{j,k}$ is re-computed accordingly. To avoid the redundant selection of information, we mask the fine-grained reliability scores of triples already included in the existing reasoning chains $\mathcal{C}^{i-1}$ with an extremely large negative value before ranking. For each query $q^i_j$, this penalization ensures that the subsequent selection prioritizes unexplored information. Accordingly, we denote $\mathcal{S}^i_j$ as the set of top-$K$ most reliable triples selected from the set $\mathcal{T}$ for each query $q^i_j$. Furthermore, these triples are both semantically relevant to $q^i_j$ and credible, serves as a highly trustworthy search space to facilitate the construction of the reasoning chains $\mathcal{C}^i$ at hop $i$. Experimental results in Section~\ref{eff_mis_sup} demonstrate that the dual-factor perception mechanism effectively filters out deceptive misinformation by identifying reliable triples. This mechanism plays a crucial role in providing effective guidance for constructing reasoning chains, thereby significantly enhancing the ReliableRAG's overall robustness and final answer accuracy.

\subsection{Chain Construction}
The objective of this module is to autoregressively construct robust reasoning chains by adopting a beam search strategy, where at each hop $i$, the chain for each query $q^i_j$ is progressively constructed by utilizing its respective set of top-$K$ most reliable triples $\mathcal{S}^i_j$. 
Unlike the prior study \cite{RF29} that are susceptible to deceptive misinformation during chain construction, our approach uniquely prioritizes and leverages reliable triples to build robust reasoning chains. Specifically, we employ a multiple-option reasoning prompt to guide the LLM-based selector $\mathcal{M}_{\text{sel}}$ in identifying the top-$B$ most probable triples from the set $\mathcal{S}^i_j$ to construct multiple candidate reasoning chains (experiments on different selectors are provided in Appendix~\ref{apx:sel}). Formally, at each hop $i$, the reasoning prompt is formulated as follows: \begin{align}
I^i_j = \Psi(\text{Instr} , \text{Exemp} , q^i_j, \mathcal{S}^i_j)
\end{align} where $\text{Instr}$ and $\text{Exemp}$ denote the task instruction and the three in-context exemplars most semantically similar to the current question $Q$, selected from a manually annotated set, respectively. Following the prior study\cite{RF29}, these exemplars are curated to maintain no overlap with the test datasets. To construct the prompt $I^i_j$, the function $\Psi(\cdot)$ first incorporates a specific task instruction and the selected exemplars into the context to guide the reasoning process. Subsequently, it organizes the termination option and the set of reliable triples $\mathcal{S}^i_j$ into an enumerated list. In this list, the termination option is assigned to option A, indicating that the current reasoning chain $u^{i-1}_j \in \mathcal{C}^{i-1}$ is sufficient to answer the question $Q$. The triples in $\mathcal{S}^i_j$ are then sequentially mapped to the subsequent option labels, such as B, C, D, and so forth, to form the complete candidate options. Guided by the reasoning prompt $I^i_j$, the LLM-based selector $\mathcal{M}_{\text{sel}}$ identifies the label corresponding to the termination signal or the next possible triple that best facilitates answering the question $Q$. By extracting the probability distribution over all option labels from $\mathcal{M}_{\text{sel}}$, we select top-$B$ highest-probability options to make multiple next-hop decisions. This selection is formulated as follows: \begin{align}
\mathcal{E}^i_j = \operatorname{Top-}B_{e_k} \Big( P_{\mathcal{M}_{\text{sel}}}(e_k \mid I^i_j) \Big)
\end{align} where $\mathcal{E}^i_j$ denotes the set of the $B$ highest-probability options selected according to the probability distribution. Subsequently, for each selected option $e_k \in \mathcal{E}^i_j$, the previous reasoning chain $u^{i-1}_j$ undergoes branching to form a candidate reasoning chain for current hop $i$, contingent on the type of the selected option. Specifically, if $e_k$ represents the termination signal, the chain $u^{i-1}_j$ is designated as a candidate reasoning chain for current hop $i$ without further expansion. Otherwise, it is expanded by appending the triple $t_k \in \mathcal{S}^i_j$ corresponding to the chosen option label. This branching process uses the beam search strategy to allow for the parallel exploration of multiple promising candidate chains, and is formally defined as: \begin{align}
u^i_m =
\begin{cases}
u^{i-1}_j, & e_k \text{ is option A} \text{ or } u^{i-1}_j \text{ is already terminated}, \\
u^{i-1}_j \oplus t_k, & \text{otherwise},
\end{cases}
\end{align} where $u^i_m$ denotes a candidate reasoning chain generated at hop $i$ representing a branch derived from the previous chain $u^{i-1}_j$ and the selected option $e_k \in \mathcal{E}^i_j $. The operator $\oplus$ denotes the concatenation of the new triple $t_k$ to the existing sequence $u^{i-1}_j$. Let $\rho^i_{j,k}$ denote the selection probability of $e_k$ for $u^{i-1}_j$, and the confidence of this branch is then measured by its cumulative probability, computed as follows: \begin{align}
\omega^i_m = \omega^{i-1}_j \cdot \rho^i_{j,k}
\end{align} where $\omega^{i-1}_j$ is the confidence of the previous chain $u^{i-1}_j$. Starting from the set of reasoning chains $\mathcal{C}^{i-1}$, all candidate reasoning chains collected for the current hop $i$ are ranked by their confidence, with only the top-$F$ most confident candidate chains retained to form $\mathcal{C}^i$ at hop $i$. This autoregressive construction process continues until the maximum hop $i = L$ is reached or all reasoning chains in the set have reached the termination signal. At the final hop $L$, which defines the maximum number of triples allowed in each chain, this module constructs the comprehensive and robust set of reasoning chains $\mathcal{C}_{\text{final}}$. Consequently, the sequence of triples within each reasoning chain in $\mathcal{C}_{\text{final}}$ provides diverse and reliable evidence to derive the answer to the question $Q$. Ablation studies in Section~\ref{exp:abl_sty} demonstrate that this module is indispensable to the overall performance of ReliableRAG. The complete procedure of the Reliability-Guided Chain Construction is formally detailed in Algorithm~\ref{alg:1}.

\begin{algorithm}
\caption{Reliability-Guided Chain Construction}
\label{alg:1}
\begin{tabbing}
\textbf{Input:} \=  question $Q$, triples with injected misinformation \\$\mathcal{T}_{\text{mis}}$, maximum number of reasoning hops $L$, number of reasoning \\chains to be retained $F$, beam size $B$, number of selected triples $K$ \\
\textbf{output:} Reasoning Chains $\mathcal{C}_{\text{final}}$
\end{tabbing}
\begin{algorithmic}[1]
\State initialize reasoning chains $\mathcal{C}^0 \gets \emptyset$ and confidence score for each chain $\Omega^0 \gets \{1.0\}$
\For {$i=1$ to $L$}
    \State Generate queries $\{q^i_j\}_{j=1}^{|\mathcal{C}^{i-1}|}$ from $Q$ and $\mathcal{C}^{i-1}$ {\color{gray} \# $|\mathcal{C}^{i-1}| \le F$}
    \State Obtain embeddings $\{h^i_j\}_{j=1}^{|\mathcal{C}^{i-1}|}$ and $\{h_k\}_{k=1}^M$ 
    \State Compute similarity $\Phi(h^i_j, h_k)$ for $j \in [1, {|\mathcal{C}^{i-1}|}\}], k \in [1, M]$
    \State Obtain credibility score $\eta_k$ for each triple
    \State Compute reliability score $R^i_{j,k} \gets \alpha \cdot \Phi(h^i_j, h_k) + (1 - \alpha) \cdot \eta_k$
    \State Mask reliability score $R^i_{j,k} \gets -\infty, \forall t_k \in u^{i-1}_j, u^{i-1}_j \in \mathcal{C}^{i-1}$
    \State {\color{gray} \# Ensuring non-redundant triple selection}
    \State Select top-$K$ most reliable triples $\mathcal{S}^i_j$ by reliability
    \State Construct reasoning prompt $I^i_j \gets \Psi(\text{Instr} , \text{Exemp} , q^i_j, \mathcal{S}^i_j)$
    \State Identify top-$B$ likely options $\mathcal{E}^i_j \gets \operatorname{Top-}B_{e_k} ( P_{\mathcal{M}_{\text{sel}}}(e_k \mid I^i_j))$
    \For{each option $e_k$ in $\mathcal{E}^i_j$}
        \If{$e_k \text{ is option A} \text{ or } u^{i-1}_j \text{ is already terminated}$}
            \State Retain the current chain $u^i_m \gets u^{i-1}_j$
        \Else
            \State Extend the chain: $u^i_m \gets u^{i-1}_j \oplus t_k$
        \EndIf
        \State Update confidence score $\omega^i_m = \omega^{i-1}_j \cdot \rho^i_{j,k}$ 
        \State {\color{gray} \# Accumulate confidence with selection probability}
        \State Retain the top-$F$ candidate chains $\mathcal{C}^i$ by confidence
    \EndFor
\EndFor
\State \textbf{return} $\mathcal{C}_{\text{final}}$
\end{algorithmic}
\end{algorithm}

\begin{table*}[!t]
\centering
\caption{Main results between ReliableRAG and compared methods under both the ideal setting and the evaluator-generated setting. The best performance within each setting is highlighted in bold. `w/ Low-credibility Docs' denotes the injection of one low-credibility document per question. The dash `-' denotes that the method is either not applicable to the setting or identical in practice to another setting already reported.}
\label{tab:1}
\scalebox{0.9}{
\begin{tabular}{cl|cccccc|cccccc} 
\hline
\multicolumn{2}{c|}{\multirow{3}{*}{Methods}} & \multicolumn{6}{c|}{Ideal Setting} & \multicolumn{6}{c}{Evaluator-Generated Setting} \\
\cline{3-14}
& & \multicolumn{2}{c}{HotPotQA} & \multicolumn{2}{c}{2WikiMultiHopQA} & \multicolumn{2}{c|}{MuSiQue} & \multicolumn{2}{c}{HotPotQA} & \multicolumn{2}{c}{2WikiMultiHopQA} & \multicolumn{2}{c}{MuSiQue} \\
\cline{3-14}
& & EM (\%) & F1 (\%) & EM (\%) & F1 (\%) & EM (\%) & F1 (\%) & EM (\%) & F1 (\%) & EM (\%) & F1 (\%) & EM (\%) & F1 (\%) \\
\hline

\multicolumn{2}{l|}{\textbf{Fine-tuning Methods}} & \multicolumn{6}{>{\columncolor{tablegray}}c|}{\textit{w/ Low-credibility Docs}} & \multicolumn{6}{>{\columncolor{tablegray}}c}{\textit{w/ Low-credibility Docs}} \\
& CAG-7B (EMNLP'24) & 5.80 & 21.53 & 3.80 & 18.20 & 1.20 & 9.45 & 4.70 & 17.91 & 3.70 & 16.58 & 0.7 & 7.71 \\
& Self-RAG-7B (ICLR'24) & 18.30 & 28.00 & 5.80 & 10.50 & 6.60 & 15.72 & - & - & - & - & - & - \\
& Knowledge-R1-7B (arxiv'25) & 26.80 & 35.45 & 18.00 & 22.81 & 11.20 & 20.76 & - & - & - & - & - & - \\
\hline

\multicolumn{2}{l|}{\textbf{Non-fine-tuning Methods}} & \multicolumn{6}{c|}{} & \multicolumn{6}{c}{} \\[4pt]

\multirow{11}{*}{\rotatebox{90}{Gemma-7B}} 
& \multicolumn{1}{c|}{} & \multicolumn{6}{>{\columncolor{tablegray}}c|}{\textit{w/o Low-credibility Docs}} & \multicolumn{6}{>{\columncolor{tablegray}}c}{\textit{w/o Low-credibility Docs}} \\
&  Naive LLM & 16.50 & 24.42 & 20.10 & 23.96 & 2.10 & 6.50 & - & - & - & - & - & - \\
&  Vanilla RAG & 43.90 & 55.47 & 30.30 & 36.93 & 17.40 & 25.61 & - & - & - & - & - & - \\[4pt] 
& \multicolumn{1}{c|}{} & \multicolumn{6}{>{\columncolor{tablegray}}c|}{\textit{w/ Low-credibility Docs}} & \multicolumn{6}{>{\columncolor{tablegray}}c}{\textit{w/ Low-credibility Docs}} \\
& Vanilla RAG & 20.10 & 28.39 & 14.50 & 19.01 & 10.40 & 20.08 & - & - & - & - & - & - \\
& Prompt Based & 43.10 & 54.42 & 32.90 & 39.03 & 18.10 & 26.52 & 33.90 & 43.65 & 27.00 & 32.80 & 14.00 & 21.16 \\
& Exclusion & - & - & - & - & - & - & 34.10 & 43.80 & 27.00 & 32.80 & 14.00 & 21.16 \\
& CrAM (AAAI'25) & 42.70 & 54.30 & 27.30 & 32.64 & 20.40 & 28.71 & 32.00 & 40.31 & 17.80 & 22.31 & 14.00 & 23.76 \\
& TruthfulRAG (AAAI'26) & 36.70 & 48.48 & 25.70 & 31.85 & 19.50 & 29.07 & - & - & - & - & - & - \\
& ReliableRAG-TBS (ours) & 52.20 & 66.89 & \textbf{45.80} & \textbf{55.88} & 30.30 & 38.66 & \textbf{39.40} & \textbf{52.96} & \textbf{32.10} & \textbf{39.30} & \textbf{19.40} & \textbf{27.21} \\
& ReliableRAG-SBS (ours) & \textbf{53.20} & \textbf{66.98} & 43.10 & 53.90 & \textbf{31.60} & \textbf{40.51} & 31.50 & 42.27 & 24.80 & 31.77 & 17.00 & 25.56 \\
\hline

\multirow{11}{*}{\rotatebox{90}{Llama3-8B}} 
& \multicolumn{1}{c|}{} & \multicolumn{6}{>{\columncolor{tablegray}}c|}{\textit{w/o Low-credibility Docs}} & \multicolumn{6}{>{\columncolor{tablegray}}c}{\textit{w/o Low-credibility Docs}} \\
&  Naive LLM & 18.90 & 27.03 & 16.10 & 23.16 & 2.70 & 6.94 & - & - & - & - & - & - \\
&  Vanilla RAG & 49.90 & 64.38 & 23.50 & 43.33 & 18.90 & 26.71 & - & - & - & - & - & - \\[4pt] 
& \multicolumn{1}{c|}{} & \multicolumn{6}{>{\columncolor{tablegray}}c|}{\textit{w/ Low-credibility Docs}} & \multicolumn{6}{>{\columncolor{tablegray}}c}{\textit{w/ Low-credibility Docs}} \\
&  Vanilla RAG & 19.30 & 28.12 & 12.50 & 19.93 & 3.30 & 12.82 & - & - & - & - & - & - \\
&  Prompt Based & 48.40 & 62.98 & 30.70 & 45.06 & 20.10 & 27.47 & 38.50 & 51.30 & 23.40 & 35.36 & 15.30 & 22.00 \\
&  Exclusion & - & - & - & - & - & - & 38.60 & 51.40 & 23.40 & 35.36 & 15.30 & 22.00 \\
&  CrAM (AAAI'25) & 50.80 & 64.43 & 20.00 & 39.64 & 19.80 & 27.48 & 33.40 & 44.21 & 19.20 & 33.81 & 8.30 & 15.93 \\
&  TruthfulRAG (AAAI'26) & 38.10 & 49.61 & 28.00 & 35.39 & 16.90 & 25.48 & - & - & - & - & - & - \\
&  ReliableRAG-TBS (ours) & 53.40 & 67.02 & \textbf{45.40} & 56.19 & 29.10 & 36.73 & \textbf{40.30} & \textbf{52.45} & \textbf{29.90} & \textbf{38.45} & \textbf{19.50} & \textbf{25.43} \\
&  ReliableRAG-SBS (ours) & \textbf{55.10} & \textbf{68.80} & 43.70 & \textbf{59.67} & \textbf{30.90} & \textbf{38.55} & 32.30 & 42.93 & 23.60 & 34.00 & 17.60 & 24.84 \\
\hline

\multirow{11}{*}{\rotatebox{90}{Mistral-7B}} 
& \multicolumn{1}{c|}{} & \multicolumn{6}{>{\columncolor{tablegray}}c|}{\textit{w/o Low-credibility Docs}} & \multicolumn{6}{>{\columncolor{tablegray}}c}{\textit{w/o Low-credibility Docs}} \\
&  Naive LLM & 19.60 & 27.22 & 22.50 & 26.41 & 3.40 & 8.02 & - & - & - & - & - & - \\
&  Vanilla RAG & 36.30 & 49.41 & 29.10 & 36.19 & 15.00 & 21.89 & - & - & - & - & - & - \\[4pt] 
& \multicolumn{1}{c|}{} & \multicolumn{6}{>{\columncolor{tablegray}}c|}{\textit{w/ Low-credibility Docs}} & \multicolumn{6}{>{\columncolor{tablegray}}c}{\textit{w/ Low-credibility Docs}} \\
&  Vanilla RAG & 18.50 & 26.05 & 11.20 & 15.71 & 8.20 & 17.55 & - & - & - & - & - & - \\
&  Prompt Based & 45.10 & 56.71 & 32.00 & 38.87 & 16.40 & 23.86 & 35.10 & 44.89 & 24.40 & 30.34 & 10.20 & 17.80 \\
&  Exclusion & - & - & - & - & - & - & 35.30 & 45.16 & 24.40 & 30.34 & 10.20 & 17.80 \\
&  CrAM (AAAI'25) & 37.20 & 49.66 & 24.60 & 32.88 & 19.50 & 29.13 & 29.00 & 38.39 & 17.50 & 23.05 & 12.70 & 21.90 \\
&  TruthfulRAG (AAAI'26) & 34.10 & 45.95 & 26.30 & 31.89 & 17.10 & 26.50 & - & - & - & - & - & - \\
&  ReliableRAG-TBS (ours) & 52.20 & 65.22 & \textbf{49.50} & \textbf{57.99} & 28.90 & \textbf{37.97} & \textbf{40.10} & \textbf{51.84} & \textbf{33.80} & \textbf{40.13} & \textbf{18.30} & \textbf{25.93} \\
&  ReliableRAG-SBS (ours) & \textbf{53.30} & \textbf{66.61} & 43.60 & 53.36 & \textbf{29.30} & 37.94 & 33.90 & 44.10 & 24.70 & 31.92 & 15.60 & 24.31 \\
\hline 
\end{tabular}
}
\end{table*}

\section{Experiments}
To comprehensively evaluate ReliableRAG, we first conduct extensive experiments on three multi-hop QA datasets. Next, we conduct ablation studies to validate the effectiveness of each individual module. Finally, we perform sensitivity analyses to investigate the impact of key hyperparameters. Detailed experimental setup, the two synthesis strategies (TBS and SBS), and an in-depth case study are presented in Appendices~\ref{apx:detail_es} to \ref{apx:detailed_case}, respectively. Additional results including experiments under the evaluator-generated setting, additional analysis (including the ablation study for ReliableRAG-SBS), experiments on different extractor, evaluator, and selector, and analysis of evaluator capability are available in Appendices~\ref{apx:model_generated} to \ref{apx:any_m_gen}.

\subsection{Experimental Setups}
\subsubsection{Datasets.} Experiments are conducted on three multi-hop QA datasets: HotPotQA \cite{RF24}, 2WikiMultiHopQA \cite{RF25}, and MuSiQue \cite{RF26}. These datasets typically require 2--4 reasoning hops, with each question associated with 10, 10, and 20 Wikipedia-sourced documents, respectively. Following the preprocessing by Fang et al. \cite{RF29}, we filter out sensitive content. To balance computational efficiency with statistical significance, we randomly sample 1,000 questions from each test set for evaluation, while an additional 100 questions from each development set are utilized for hyperparameter tuning. Misinformation scenarios are simulated by injecting low-credibility documents (each containing LLM-generated deceptive misinformation) for each question, with performance evaluated under both Ideal and Evaluator-generated settings.

\begin{figure}[!t]
\centerline{\includegraphics[width=\columnwidth]{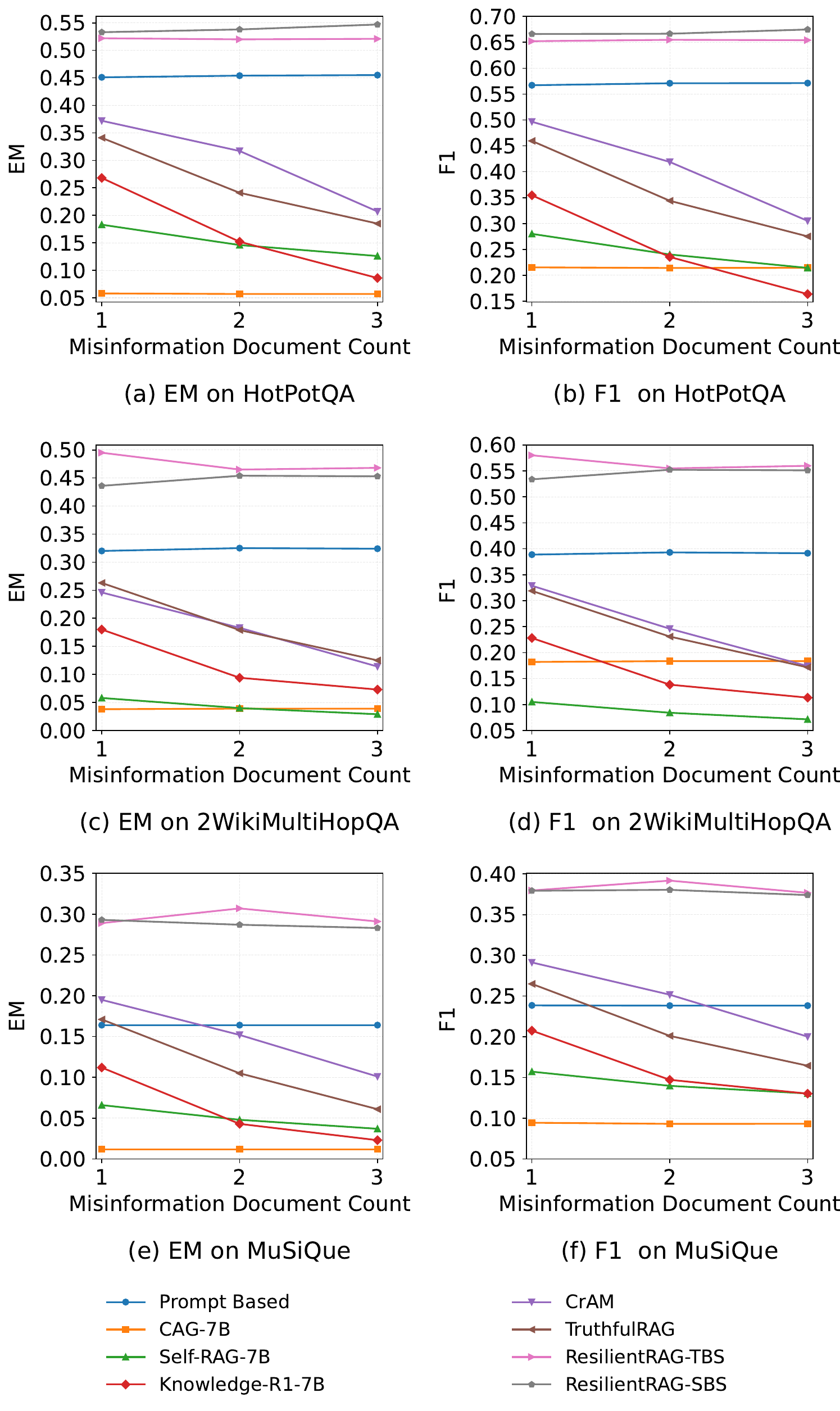}}
\caption{Performance comparison between ReliableRAG and six compared methods on the test sets of three multi-hop QA datasets with varying numbers of low-credibility documents per question, evaluated under the ideal setting.}
\label{fig:5}
\end{figure}
\subsubsection{Compared Methods and Evaluation Metrics.} ReliableRAG is compared against a diverse set of compared methods, including Naive LLM, Vanilla RAG, Prompt-based, Exclusion, Self-RAG \cite{RF68}, CAG \cite{RF11}, Knowledgeable-R1 \cite{RF67}, CrAM \cite{RF22}, TruthfulRAG \cite{RF66}. To comprehensively evaluate the performance of all methods, we adopt two standard QA metrics: Exact Match (EM) \cite{RF58,RF59} and F1 score (F1) \cite{RF56,RF57}. 


\subsection{Main Results}
\subsubsection{Performance Comparison}
Table~\ref{tab:1} presents the comparison between ReliableRAG and nine compared methods, covering both fine-tuning and non-fine-tuning methods. The results clearly demonstrate that ReliableRAG significantly outperforms all compared methods under the Ideal setting. Across the three tested LLMs, it delivers the highest EM performance on all test sets, achieving 55.10\%, 49.50\%, and 31.60\% on HotPotQA, 2WikiMultiHopQA, and MuSiQue, respectively. Similarly, ReliableRAG attains the highest EM performance under the evaluator-generated setting across all tested LLMs, yielding 40.30\%, 33.80\%, and 19.50\% on HotPotQA, 2WikiMultiHopQA, and MuSiQue, respectively. These findings underscore the superior robustness of ReliableRAG in resisting the interference of deceptive misinformation, preventing its propagation during multi-hop reasoning, and maintaining stability during complex multi-hop QA tasks. Notably, the CAG method exhibits significant instruction-following difficulties and tends to generate excessive explanatory content, which leads to abysmal EM performance.


\subsubsection{Robustness to Misinformation}
\label{Impact_mis_docs}
To further validate robustness, we vary the number of low-credibility documents per question on all test sets, considering both ideal and evaluator-generated settings using Mistral-7B as the generator. As shown in Figure~\ref{fig:5}, ReliableRAG outperforms six compared methods with minimal performance drops. Results in evaluator-generated settings exhibit a similar pattern (see Appendix~\ref{apx:Detailed_pc} for detailed results). Interestingly, we observe that the EM performance of ReliableRAG-SBS on HotpotQA even improves slightly as the misinformation proportion rises. We hypothesize that our framework can uncover fine-grained correct information hidden even within low-credibility documents, subsequently transforming these fine-grained informational into structured triples that are effectively utilized to derive the final answer. We attribute this capability to the dual-factor perception mechanism within the Triple Evaluation module, which effectively captures these correct triples and integrates them into robust reasoning chains. This allows the framework to leverage valid fine-grained information while resisting deceptive misinformation, thereby enhancing the framework's overall robustness. Furthermore, our framework ultimately yields a concise input context $\mathcal{X}$ for final generation, the input computational overhead of which is detailed in Appendix~\ref{apx:Context_Length} through experimental results.

\subsection{Ablation Studies}
\label{exp:abl_sty}
To evaluate the contribution of each component within ReliableRAG-TBS, we conduct a systematic ablation study. For the `w/o Chain Construction' variant, instead of generating a reasoning chain, we employ E5-Mistral to retrieve the top-$K$ most reliable triples directly from the set of triples $\mathcal{T}$ for each query. These triplets are then provided as the supporting context for the generator $\mathcal{M}_{\text{gen}}$ to produce the final answer. To ensure a fair comparison, we evaluate this variant across various values of $K$ and report the corresponding results.
\par
Results in Table~\ref{tab:2} show that the complete ReliableRAG-TBS achieves best performance across all test sets (Detailed analysis for ReliableRAG-SBS is provided in Appendix~\ref{apx:Detailed ablation study for ReliableRAG-SBS}). Notably, removing Triple Extraction triggers the most substantial performance decline, represented by a 34.1\% drop in EM on the HotpotQA test set, underscoring the triples' pivotal role in organizing information for multi-hop QA. Similarly, excluding Triple Evaluation leads to a 15.6\% EM decrease, highlighting its importance in preventing misinformation propagation during multi-hop reasoning. Furthermore, we observe a distinct performance threshold in the ablation variant without Chain Construction. Specifically, the EM performance of ReliableRAG-TBS on HotPotQA and MuSiQue begins to decline once the number of triples exceeds a certain threshold, typically between 15 and 25. This suggests that blindly increasing the number of triples does not yield further performance gains. This demonstrates the importance of Chain Construction in utilizing these triples to construct reasoning chains. Notably, while the removal of Chain Construction leads to a performance drop, the margin of decline remains relatively moderate compared to the preceding two modules. This indicates that the Triple Extraction and Triple Evaluation modules establish a search space comprising highly reliable triples, which is pivotal in preventing the propagation of misinformation during multi-hop reasoning. Building upon this, the Chain Construction module constructs robust multi-hop reasoning chains from the highly reliable triples within the search space. These chains aggregate as many logically coherent and reliable triples as possible, ultimately synthesizing them into a comprehensive and reliable supporting context, thereby further fortifying the system's overall robustness and factual reliability, and ensuring final answer precision.

\begin{table}[!t]
\centering
\caption{Ablation results of ReliableRAG-TBS on the test sets of three multi-hop QA datasets. The Top-$K$ Triple variants represent the performance where the supporting context $\mathcal{Z}$ is constructed by applying the TBS synthesis strategy to the top-$K$ most reliable triples without the Chain Construction module.}
\label{tab:2}
\small 
\setlength{\tabcolsep}{2.5pt} 
\scalebox{0.9}{
\begin{tabular}{lcccccc}
\toprule
\multirow{2}{*}{\textbf{Methods}} & \multicolumn{2}{c}{\textbf{HotPotQA}} & \multicolumn{2}{c}{\textbf{2WikiMultiHopQA}} & \multicolumn{2}{c}{\textbf{MuSiQue}} \\ \cmidrule(lr){2-3} \cmidrule(lr){4-5} \cmidrule(lr){6-7} 
 & EM (\%) & F1 (\%) & EM (\%) & F1 (\%) & EM (\%) & F1 (\%) \\ \midrule
w/o Triple Extraction & 19.30 & 28.12 & 12.50 & 19.93 & 3.30 & 12.82 \\ \midrule
w/o Triple Evaluation  & 37.80 & 49.42 & 30.10 & 37.16 & 17.20 & 25.81 \\ \midrule
\textit{w/o Chain Construction} & & & & & & \\
\quad Top-10 Triples & 46.40 & 59.84 & 26.10 & 37.53 & 21.30 & 28.61 \\
\quad Top-15 Triples & 46.40 & 60.01 & 29.70 & 42.39 & 23.90 & 31.96 \\
\quad Top-20 Triples & 47.50 & 61.37 & 30.80 & 45.30 & 22.80 & 31.17 \\
\quad Top-25 Triples & 47.00 & 60.98 & 32.20 & 48.38 & 23.10 & 31.67 \\ \midrule
ReliableRAG-TBS & \textbf{53.40} & \textbf{67.02} & \textbf{45.40} & \textbf{56.19} & \textbf{29.10} & \textbf{36.73} \\ \bottomrule
\end{tabular}
}
\end{table}

\begin{figure}[!t]
\centerline{\includegraphics[width=\columnwidth]{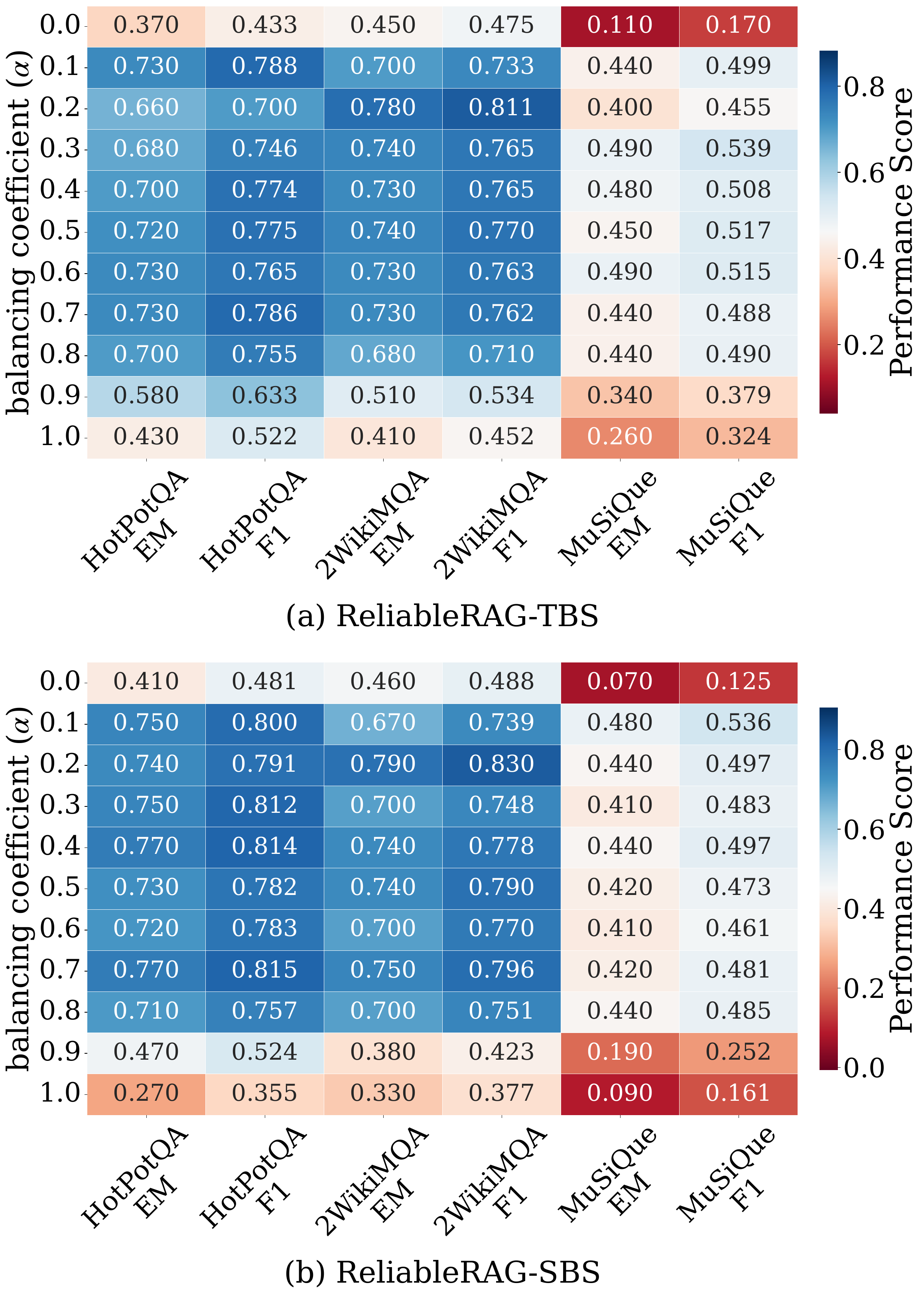}}
\caption{EM and F1 performance of ReliableRAG across various configurations of the balancing coefficient $\alpha$ that regulates the dual-factor perception mechanism within the Triple Evaluation module. Results compare (a) ReliableRAG-TBS and (b) ReliableRAG-SBS variants, evaluated under the ideal setting using the dev sets of three multi-hop QA datasets.}
\label{fig:6}
\end{figure}

\subsection{Sensitivity analyses}
\subsubsection{Trade-off in Triple Evaluation}
We systematically evaluate the influence of the balancing coefficient $\alpha$ that regulates the dual-factor perception mechanism within the Triple Evaluation module. Using Llama3-8B as the generator, we perform a grid search for $\alpha \in [0, 1.0]$ across the development sets of three multi-hop QA datasets, each injected with one low-credibility document per question. As illustrated in Figure~\ref{fig:6}, ReliableRAG achieves its most balanced performance at $\alpha^*=0.4$, striking an optimal trade-off between query-triple semantic relevance and triple credibility. Specifically, while a larger $\alpha$ prioritizes semantic relevance, it renders the framework more susceptible to deceptive misinformation. Conversely, a smaller $\alpha$ more effectively suppresses misinformation but may overly restrict information utilization, thereby compromising the overall generation accuracy.

\subsubsection{Effectiveness in Preventing Misinformation}
\label{eff_mis_sup}
\begin{figure}[htbp]
\centerline{\includegraphics[width=\columnwidth]{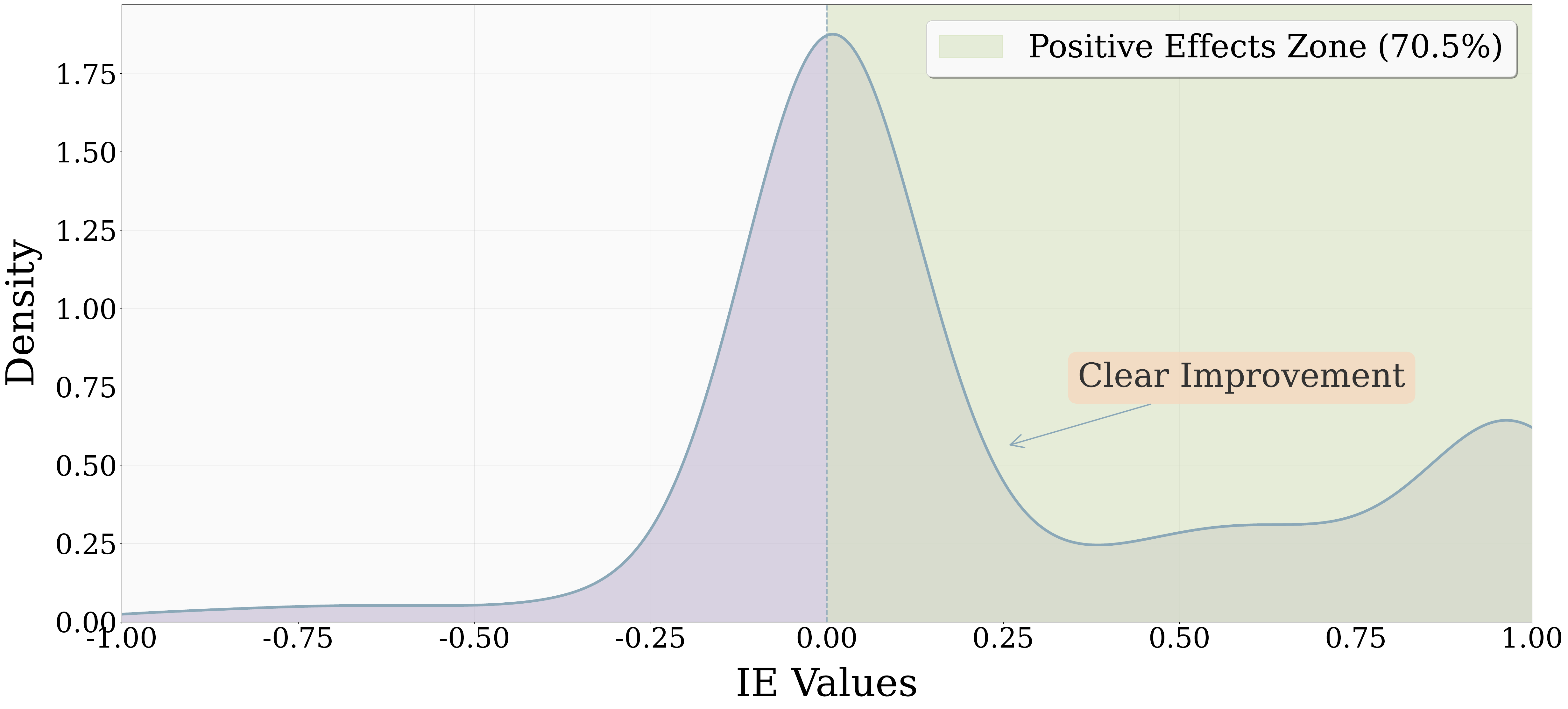}}
\caption{IE density distribution under the optimal configuration $\alpha^*=0.4$.}
\label{fig:7}
\end{figure}

Following the prior study \cite{RF22} that quantifies the contribution of specific components by measuring the change in generation probability, we adopt the Indirect Effect (IE) to validate the efficacy of the dual-factor perception mechanism within the Triple Evaluation module and its role in guiding the construction of reasoning chains. The formal definition of IE is detailed in Appendix~\ref{apx:def_ie}. Specifically, across the development sets of three multi-hop QA datasets (each injected with one low-credibility document per question), we analyze the distribution of the IE for all ReliableRAG variants (including both TBS and SBS) at the optimal balancing coefficient $\alpha^*=0.4$. As illustrated in Figure~\ref{fig:7}, the distribution approximates a normal shape centered near zero, with a significant density shift toward the positive side. Notably, positive effects account for 70.5\% of the cases, indicating that our balancing coefficient adjustment generally enhances reasoning correctness. These findings confirm that the dual-factor perception mechanism not only fortifies ReliableRAG against misinformation but also provides effective guidance for constructing reasoning chains, thereby bolstering the overall robustness and factual reliability of the RAG system.

\subsubsection{Impact of Reasoning Depth}
\begin{figure}[htbp]
\centerline{\includegraphics[width=\columnwidth]{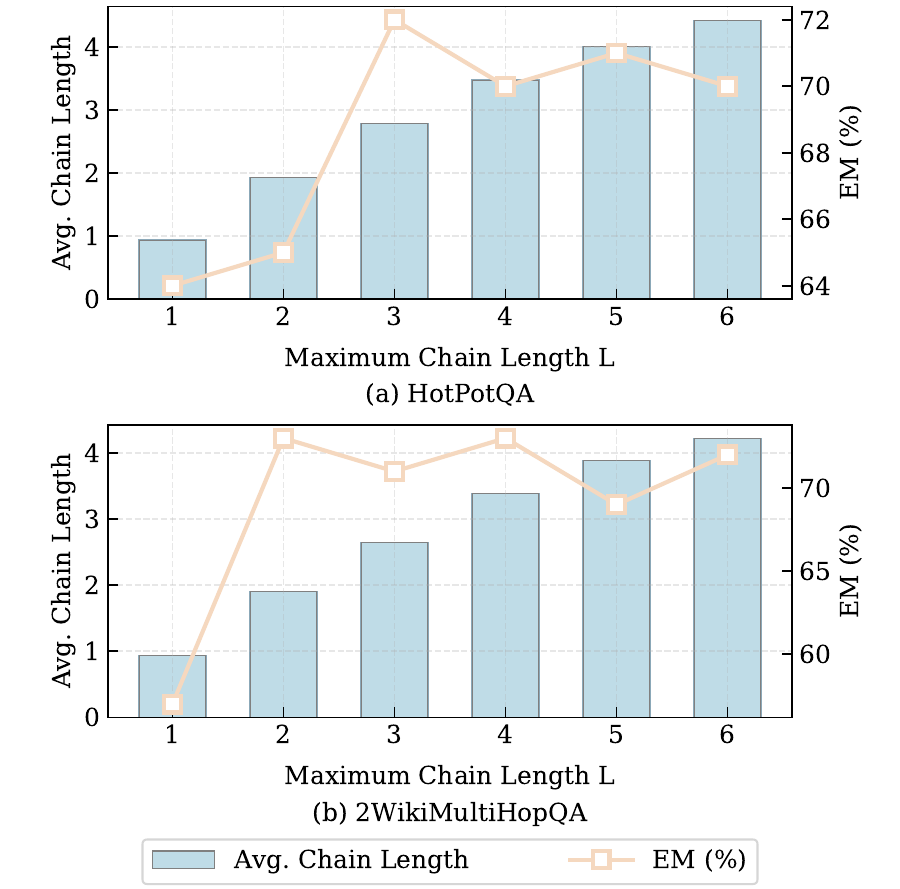}}
\caption{Impact of the maximum reasoning chain length $L$ on the QA performance and average chain length of ReliableRAG-TBS on the dev sets of two multi-hop QA datasets, under an ideal setting where one low-credibility document is injected per question.}
\label{fig:8}
\end{figure}

To investigate how the maximum allowable number of reasoning steps (i.e., the maximum reasoning chain length $L$) affects performance, we perform a sensitivity analysis of ReliableRAG by varying $L$ from 1 to 6. As illustrated in Figure~\ref{fig:8}, across two dev sets, the average chain length exhibits sub-linear growth as $L$ increases, eventually plateauing rather than increasing indefinitely. We observe that although the average chain length continues to grow albeit at a much slower rate as $L$ increases from 4 to 6, this additional depth does not translate into detectable EM gains. Notably, when $L > 4$, the EM performance remains virtually stagnant or even experiences a slight decline (additional results on the MuSiQue dataset are detailed in Appendix~\ref{apx:Results_on_MuSiQue}). This suggests that excessively increasing the reasoning depth does not yield further gains. Instead, it may introduce deceptive misinformation that contaminates the reasoning process, ultimately undermining the system's robustness.

\section{Conclusion}
In this paper, we introduce ReliableRAG, a novel framework designed to combat deceptive misinformation in multi-hop QA. By dynamically evaluating fine-grained reliability at the triple level, our approach filters deceptive misinformation to construct reliable, autoregressive reasoning chains. Extensive experiments demonstrate that ReliableRAG significantly outperforms existing methods, providing RAG systems with a highly robust solution to ensure factual reliability and prevent misinformation propagation during multi-hop reasoning processes.

\begin{acks}
To Robert, for the bagels and explaining CMYK and color spaces.
\end{acks}

\clearpage
\bibliographystyle{ACM-Reference-Format}
\bibliography{ref}
\clearpage

\appendix

\section{Detailed Experimental Setup}
In this section, we provide the detailed configurations for compared methods, the document injection process, and our implementation.
\label{apx:detail_es}
\subsection{Compared Methods}
ReliableRAG is compared against a diverse set of compared methods: (1) Naive LLM, which generates answers using only internal knowledge without retrieval; (2) Vanilla RAG, which follows the standard RAG pipeline but lacks any mechanism to mitigate misinformation; (3) Prompt-based, which incorporates document credibility into prompts to guide the generation of the generator $\mathcal{M}_{\text{gen}}$ without parameter updates; (4) Exclusion, a strategy built upon prompt-based method to filter out documents below a fixed credibility threshold; (5) Self-RAG \cite{RF68}, fine-tuned on Llama2-7B that utilizes reflection tokens for self-assessment of retrieval and generation; (6) CAG \cite{RF11}, fine-tuned on Llama2-7B by encoding both content and credibility scores; (7) Knowledgeable-R1 \cite{RF67}, fine-tuned on Qwen2.5-7B-Instruct to enhance resistance against misleading contexts; (8) CrAM \cite{RF22}, which dynamically adjusts attention weights based on document credibility; and (9) TruthfulRAG \cite{RF66}, which employs knowledge graphs and entropy-based filtering to eliminate factual inconsistencies. In addition, we conduct comparative experiments by employing Llama2-7B and Qwen2.5-7B-Instruct as the generator $\mathcal{M}_{\text{gen}}$, providing a fair comparison against methods including Self-RAG, CAG, and Knowledgeable-R1 to further validate the versatility of our framework (see Appendix~\ref{extra_rlt}).

\subsection{Low-credibility Document Injection} To evaluate the effectiveness of ReliableRAG in preventing misinformation, we construct low-credibility documents that contain misinformation and support incorrect answers using an LLM (i.e., GLM-4-Flash). Following the prior study \cite{RF22}, the LLM is guided by specific prompts to generate coherent yet misleading content. For each question, we inject three diverse low-credibility documents supporting the same erroneous answer to simulate a challenging misinformation environment. Two evaluation settings are considered: (1) Ideal setting, where high-credibility (Wikipedia-sourced) and low-credibility documents are assigned fixed scores of 10 and 1, respectively, to establish an upper-bound performance; and (2) Evaluator-generated setting, where the evaluator $\mathcal{M}_{\text{eva}}$ assigns credibility scores (ranging from 0 to 10) to each document.

\subsection{Implementation Details}
\label{apx:Implementation Details}
Experiments were run on a platform with Python 3.10 and CUDA 12.1.  Unless otherwise specified, we employ Llama3-8B-Instruct as the default LLM and e5-Mistral-7B-Instruct as the default bi-encoder for computing query–document and query–triple similarities throughout the framework. The platform comprised an Intel Xeon Gold 6348 CPU, 503 GB of memory, and an NVIDIA A800 80GB GPU. Specifically, generation used a temperature of 1.0 with deterministic sampling. Whereas compared methods retrieve the top 5 documents by similarity, ReliableRAG selects the top 5 reasoning chains in $\mathcal{C}_{\text{final}}$ based on their confidence scores. These selected chains are then utilized to construct the supporting context $\mathcal{Z}$, which is integrated into the unified input $\mathcal{X}$. This input $\mathcal{X}$ is subsequently fed into the generator $\mathcal{M}_{\text{gen}}$ to produce the final answer.

\section{Prompting and Context Synthesis Strategies}
\label{apx:prompt_stn}
\begin{figure}[htbp]
\centerline{\includegraphics[width=\columnwidth]{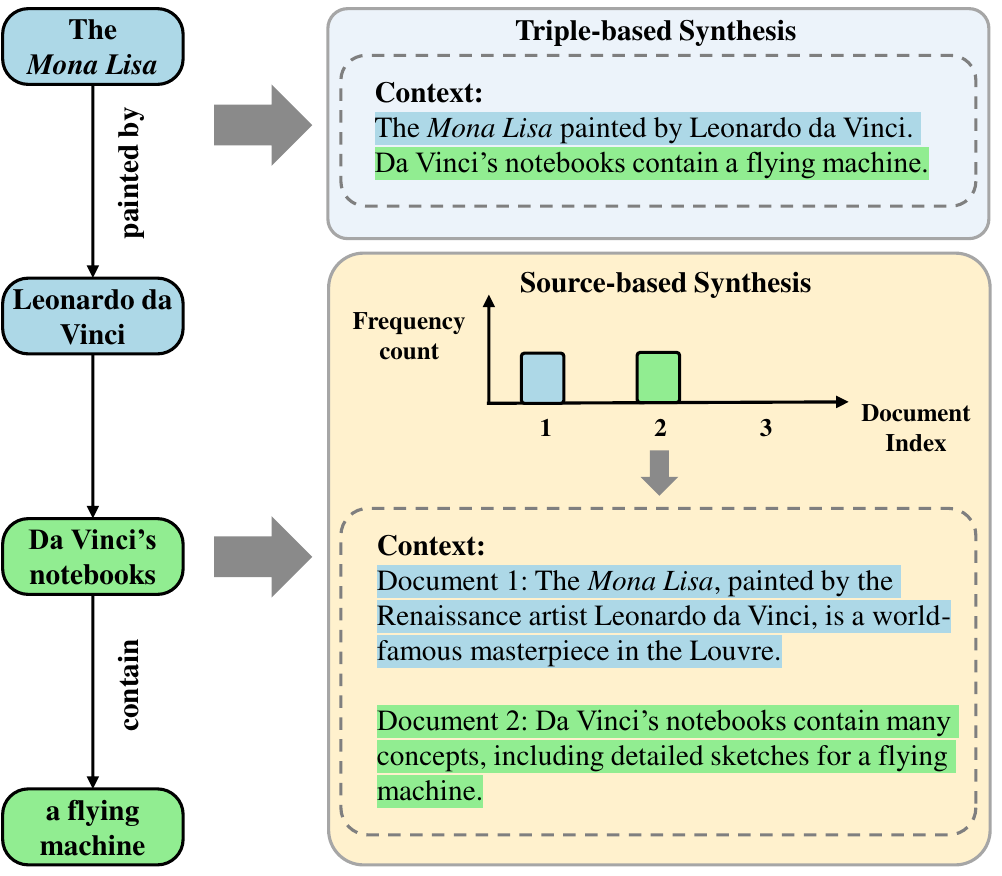}}
\caption{Illustration of the Context Synthesis Strategies. Triple-based Synthesis (TBS) directly converts triples into a concise natural language format by concatenating their components into textual strings to maximize information density. In contrast, Source-based Synthesis (SBS) utilizes the projection function $\pi(\cdot)$ to map triples back to original source documents, identifying the most representative evidence documents through a frequency-based voting mechanism, thereby restoring the rich semantic details.}
\label{fig:cxt_syns}
\end{figure}

This section provides supplementary details regarding our framework. First, the detailed content of the assessment framework is illustrated in Figure~\ref{fig:ass_frk}. Furthermore, we elaborate on the context synthesis strategies as depicted in Figure~\ref{fig:cxt_syns}. Following the construction of the reasoning chain set $\mathcal{C}_{\text{final}}$, we transform the selected chains into a reliable, evidence-based supporting context $\mathcal{Z}$ that is formatted to be comprehensible to the generator $\mathcal{M}_{\text{gen}}$. To achieve this, we employ two representative strategies to synthesize $\mathcal{Z}$:
\par
\textbf{(1) Triple-based Synthesis (TBS):}
This strategy directly converts the triples into a concise natural language format. Specifically, for each triple $t_k = (s_k, p_k, o_k)$ within each reasoning chain $c^L_j \in \mathcal{C}_{\text{final}}$, the constituent subject $s_k$, predicate $p_k$, and object $o_k$ are concatenated into a textual string. These strings are then integrated to form the supporting context $\mathcal{Z}$. By presenting the generator $\mathcal{M}_{\text{gen}}$ with this information-dense context (as part of the unified input $\mathcal{X}$), this strategy ensures that $\mathcal{M}_{\text{gen}}$ is grounded in a concise information space for final answering.

\textbf{(2) Source-based Synthesis (SBS):}
This strategy constructs the supporting context $\mathcal{Z}$ by mapping each triple back to its original source document. Specifically, for each triple $t_k$ within a reasoning chain $u^{\text{final}}_j \in \mathcal{C}_{\text{final}}$, we define a projection function $\pi(t_k)$ that maps it back to its corresponding source document $d_n \in \mathcal{D}$, thereby restoring the rich semantic details. By treating each such mapping as a vote for $d_n$, we employ a frequency-based voting mechanism to rank the documents based on their total occurrences. The documents identified through these mappings are then sequentially integrated according to their ranked order to form $\mathcal{Z}$. This ensures that the generator $\mathcal{M}_{\text{gen}}$ is grounded in the comprehensive content of the original source documents.

\section{Detailed Case Study}
\label{apx:detailed_case}
In this section, we present a case study in Table~\ref{tab:case_study} to nstantiate the internal reasoning process of ReliableRAG when encountering deceptive misinformation. This case is selected from the HotPotQA test set. The execution flow and logical evolution of each module are described as follows:
\par
\textbf{Step 1: Triple Extraction.} This module transforms the source documents into a set of structured triples. As shown in the Table~\ref{tab:case_study}, the set includes both triples from high-credibility documents (e.g., Wikipedia), such as \textit{(Leo Harris, notable deal, handshake deal with Walt Disney to use Donald Duck as the basis for the Oregon Duck mascot)}, and deceptive misinformation from low-credibility documents, such as \textit{(Mickey Mouse's Legacy in Animation Education, popular misconception, that Donald Duck was featured in the deal, not Mickey Mouse)}. By providing such fine-grained information for subsequent multi-hop reasoning, this approach effectively mitigates the lost-in-the-middle issue.
\par
\textbf{Step 2: Triple Evaluation.} Building upon the structured triples derived from the Triple Extraction module and their respective credibility scores assigned by the evaluator $\mathcal{M}_{\text{eva}}$, this module performs fine-grained reliability quantification by synthesizing query-triple semantic relevance and triple credibility. This mechanism ensures the retention of only the top-$K$ most reliable and non-redundant triples, thereby providing a highly trustworthy search space for subsequent chain construction. As illustrated in Table~\ref{tab:case_study}, this module effectively filters out triples containing deceptive misinformation while preserving the most reliable evidence triples.
\par
\textbf{Step 3: Chain Construction.} Operating on the trustworthy search space established by the Triple Evaluation module, this module employs a beam search strategy to autoregressively construct robust reasoning chains at each hop. Since the search space has been filtered for reliability, the construction process naturally avoids deceptive misinformation. As demonstrated in Table~\ref{tab:case_study}, the highest-ranked candidate chain at the second hop, which consists of \textit{(Leo Harris, notable deal, handshake deal with Walt Disney to use Donald Duck as the basis for the Oregon Duck mascot)} and \textit{(Donald Duck, creation year, 1934)}, contains sufficient information to directly answer the question $Q$.
\par
Finally, the final reasoning chains $\mathcal{C}_{\text{final}}$ are synthesized into the supporting context $\mathcal{Z}$ through context synthesis strategies, then integrated into the unified input $\mathcal{X}$ for the generator $\mathcal{M}_{\text{gen}}$. By supplying this refined and reliable context $\mathcal{Z}$, ReliableRAG effectively alleviates the burden on $\mathcal{M}_{\text{gen}}$ of integrating raw, potentially deceptive misinformation, thereby enabling the generation of the accurate final answer: ``Donald Duck.''

\section{Definition of Indirect Effect}
\label{apx:def_ie}
In this section, following the prior study \cite{RF22} that quantifies the contribution of specific components by measuring the change in generation probability, we adopt the Indirect Effect (IE) to validate the efficacy of the dual-factor perception mechanism and its role in guiding the construction of reasoning chains. Specifically, we first establish a baseline probability $P_0$, which represents the likelihood of deriving the correct answer for each question $Q$ when the balancing coefficient $\alpha$ within the dual-factor perception mechanism is set to $\alpha^0 = 1$. To compute this probability, we apply two distinct strategies to the constructed reasoning chains to synthesize two corresponding types of supporting contexts $\mathcal{Z}_k$ (where $k \in \{1, 2\}$). Each $\mathcal{Z}_k$ is independently integrated into the input $\mathcal{X}_k$ and fed to the generator $\mathcal{M}_{\text{gen}}$, yielding two separate probabilities whose combination defines $P_0$. This resulting probability $P_0$ serves as our reference standard, formulated as follows: \begin{align}
P_0=\{P_{\mathcal{M}_{\text{gen}}}(a \mid \mathcal{X}_k,\alpha^0)\}_{k=1}^2
\end{align} Subsequently, to determine the optimal balance between these factors, we conduct a grid search for $\alpha$ on the development sets of three multi-hop QA datasets, each injected with one low-credibility document per question. This process identifies the most balanced parameter configuration , denoted as $\alpha^*$, which maximizes the performance of both ReliableRAG-TBS and ReliableRAG-SBS across all tasks. The probability of generating the correct answer under this configuration is expressed as: \begin{align}
P_1=\{P_{\mathcal{M}_{\text{gen}}}(a \mid \mathcal{X}_k,\alpha^*)\}_{k=1}^2
\end{align} The influence of the balancing coefficient $\alpha$ is quantified as the IE: \begin{align}
\mathrm{IE}_{\alpha^*}=P_1-P_0
\end{align} Intuitively, the IE distribution across all evaluated questions characterizes the mechanism's overall efficacy, where a positive shift in this distribution signifies an enhanced capacity to mitigate misinformation.

\section{Robustness Analysis under Evaluator-generated Settings}
In this section, we analyze the robustness of our proposed framework under evaluator-generated settings. Specifically, we first investigate the vulnerability of compared methods, followed by a demonstration of the superior robustness of ReliableRAG.
\label{apx:model_generated}
\subsection{Vulnerability of Compared Methods}
Under the evaluator-generated setting, we conduct a comparative analysis by varying the proportion of low-credibility documents containing deceptive misinformation. As illustrated in Figure~\ref{apx:fig1}, compared methods such as CrAM and Exclusion exhibit a substantial decline in performance as the misinformation ratio increases. This pronounced degradation confirms the inherent vulnerability of these approaches when confronted with dense and deceptive misinformation.
\subsection{Robustness of ReliableRAG}
\label{apx:Detailed_pc}
\begin{figure}[!t]
\centerline{\includegraphics[width=\columnwidth]{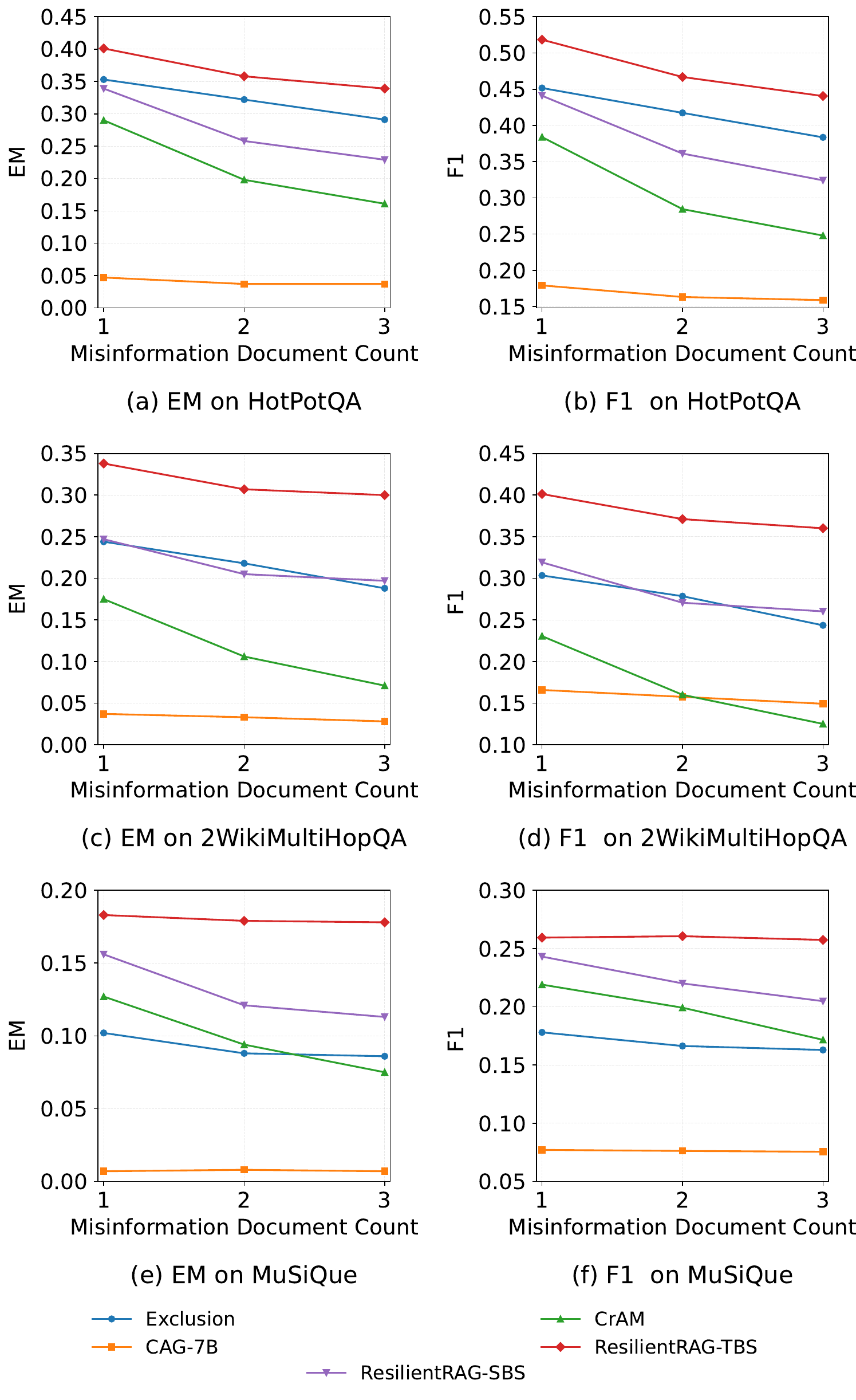}}
\caption{Performance of ReliableRAG versus compared methods under the evaluator-generated setting, across three multi-hop QA datasets with varying proportions of low-credibility documents per question.}
\label{apx:fig1}
\end{figure}

As shown in Figure~\ref{apx:fig1}, our proposed method maintains robust performance with minimal degradation, significantly outperforming the compared methods. Notably, the Exact Match (EM) performance of ReliableRAG-TBS on complex tasks such as MuSiQue remains remarkably stable even as the proportion of misinformation scales up. This observed stability potentially relates to the hypothesis proposed in Section~\ref{Impact_mis_docs}: the dual-factor perception mechanism within the Triple Evaluation module effectively captures factually correct triples within low-credibility documents while filtering out those containing misinformation. By utilizing these correct triples to construct reasoning chains, our approach successfully prevents the propagation of misinformation throughout the multi-hop reasoning process.

\section{Additional Analysis}
In this section, we provide a comprehensive analysis of various factors influencing the framework's performance. Specifically, we first investigate the sensitivity to the maximum chain length $L$ on the MuSiQue dataset, followed by an assessment of the impact of various bi-encoder models. Then, we conduct an ablation study for ReliableRAG-SBS and analyze the input computational overhead. Finally, we provide extra evidence regarding the framework's generalizability and robustness across various generators and settings.
\subsection{Sensitivity to Maximum Chain Length $L$ on MuSiQue}
\label{apx:Results_on_MuSiQue}
\begin{figure}[htbp]
\centerline{\includegraphics[width=\columnwidth]{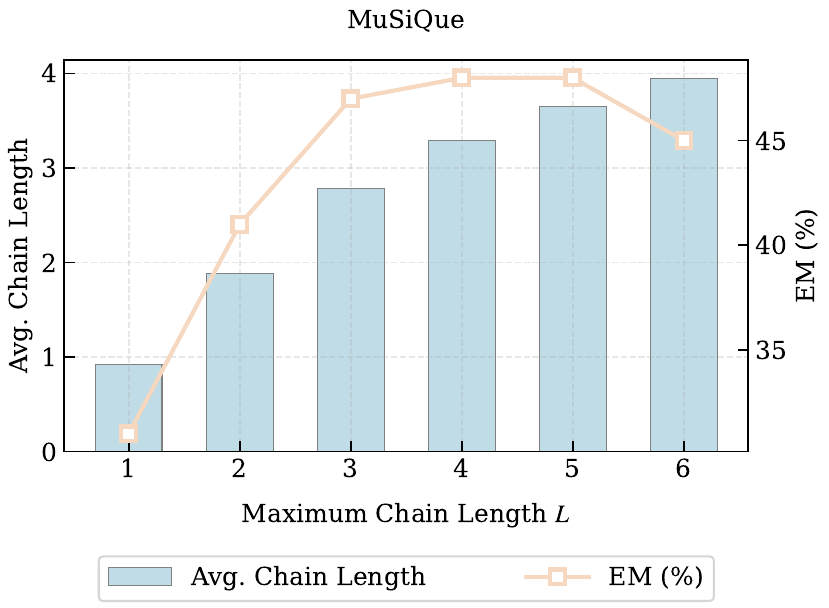}}
\caption{Impact of $L$ on the multi-hop QA performance and average chain length of ReliableRAG-TBS on the dev set of MuSiQue under the ideal setting where one low-credibility document is injected per question.}
\label{apx:fig_maxlen}
\end{figure}

As illustrated in Figure~\ref{apx:fig_maxlen}, the results on the MuSiQue development set align with our previous findings on HotpotQA and 2WikiMultiHopQA. Specifically, we observe that when $L > 4$, the growth of the average chain length slows significantly, while the EM performance remains virtually stagnant or even experiences a slight decline. This cross-dataset consistency further reinforces our earlier observation that increasing reasoning depth beyond a certain threshold yields diminishing returns, as additional reasoning hops tend to introduce deceptive misinformation that compromises overall accuracy.

\subsection{Impact of Various Bi-encoder Models}
\label{apx:bi_encoder_selection}
To determine the optimal bi-encoder for triple retrieval, we compare the performance of ReliableRAG using three representative models: DRAGON+ \cite{RF100}, E5 \cite{RF101}, and E5-Mistral \cite{RF102}. The evaluation results on the dev sets of three multi-hop QA datasets for both TBS and SBS strategies are presented in Table~\ref{tab:retrieval_comparison_1} and Table~\ref{tab:retrieval_comparison_2}. The empirical results demonstrate that E5-Mistral delivers strong performance across various datasets and strategies, particularly showing notable advantages on complex multi-hop scenarios like 2WikiMultiHopQA and MuSiQue. These findings suggest that E5-Mistral provides superior semantic alignment that effectively facilitates the subsequent reasoning chain construction. Given its robust overall performance and better adaptability to challenging datasets, we select E5-Mistral as the default bi-encoder for all primary experiments.

\begin{table}[!t]
\centering
\caption{Performance comparison of ReliableRAG-TBS employing different bi-encoder models for triple retrieval on the dev sets of three multi-hop QA datasets under the ideal setting where one low-credibility document is injected per question.}
\label{tab:retrieval_comparison_1}
\small
\begin{tabular}{lcccccc}
\toprule
\multirow{2}{*}{\textbf{Models}} & \multicolumn{2}{c}{\textbf{HotPotQA}} & \multicolumn{2}{c}{\textbf{2WikiMultiHopQA}} & \multicolumn{2}{c}{\textbf{MuSiQue}} \\ \cmidrule(lr){2-3} \cmidrule(lr){4-5} \cmidrule(lr){6-7} 
 & EM (\%) & F1 (\%) & EM (\%) & F1 (\%) & EM (\%) & F1 (\%) \\ \midrule
DRAGON+    & 70.0 & 75.3 & 66.0 & 70.0 & 44.0 & 49.9 \\
E5         & \textbf{73.0} & \textbf{79.4} & 71.0 & 73.5 & 40.0 & 46.2 \\
E5-Mistral & 70.0 & 77.4 & \textbf{73.0} & \textbf{76.5} & \textbf{48.0} & \textbf{50.8} \\ \bottomrule
\end{tabular}
\end{table}

\begin{table}[!t]
\centering
\caption{Performance comparison of ReliableRAG-SBS employing different bi-encoder models for triple retrieval on the dev sets of three multi-hop QA datasets under the ideal setting where one low-credibility document is injected per question.}
\label{tab:retrieval_comparison_2}
\small
\begin{tabular}{lcccccc}
\toprule
\multirow{2}{*}{\textbf{Models}} & \multicolumn{2}{c}{\textbf{HotPotQA}} & \multicolumn{2}{c}{\textbf{2WikiMultiHopQA}} & \multicolumn{2}{c}{\textbf{MuSiQue}} \\ \cmidrule(lr){2-3} \cmidrule(lr){4-5} \cmidrule(lr){6-7} 
 & EM (\%) & F1 (\%) & EM (\%) & F1 (\%) & EM (\%) & F1 (\%) \\ \midrule
DRAGON+    & 75.0 & 79.7 & 61.0 & 68.2 & 43.0 & 47.8 \\
E5         & 76.0 & \textbf{81.7} & 70.0 & 74.4 & 39.0 & 46.2 \\
E5-Mistral & \textbf{77.0} & 81.4 & \textbf{74.0} & \textbf{77.8} & \textbf{44.0} & \textbf{49.7} \\ \bottomrule
\end{tabular}
\end{table}

\subsection{Ablation study for ReliableRAG-SBS}
\label{apx:Detailed ablation study for ReliableRAG-SBS}
Similar to the experimental setup in Section~\ref{exp:abl_sty}, we conduct an ablation study on ReliableRAG-SBS. Results in Table~\ref{tab:apx_ablation_study} demonstrate that the complete ReliableRAG-SBS outperforms all of its variants across all test sets. Consistent with the trends observed in ReliableRAG-TBS, the removal of Triple Extraction on the HotpotQA test set triggers the most substantial performance drop of 35.8\% in EM, confirming the triples' pivotal role in organizing information for multi-hop QA.
\par
Furthermore, excluding Triple Evaluation results in a 31.0\% decrease in EM. Notably, the performance penalty for removing this stage is significantly more pronounced in the SBS variant than in TBS. This is primarily because the SBS strategy constructs the supporting context $\mathcal{Z}$ by mapping each triple back to its original coarse-grained source document. Without the Triple Evaluation process, the quality of $\mathcal{Z}$ diminishes sharply, which directly impacts ReliableRAG's ability to resist deceptive misinformation. Finally, ReliableRAG-SBS exhibits a distinct performance threshold similar to that observed in ReliableRAG-TBS. Specifically, without Chain Construction, the EM performance across all datasets begins to decline once the number of triples exceeds a certain threshold (typically between 15 and 20). This again confirms that blindly increasing the number of triples does not yield further performance gains, highlighting the critical role of Chain Construction in organizing these triples into coherent reasoning chains. Collectively, these findings further validate the efficacy of each module within our framework and its overall capacity to bolster the robustness and factual reliability of RAG systems.

\begin{table}[!t]
\centering
\caption{Ablation results of ReliableRAG-SBS on the test sets of three multi-hop QA datasets. The Top-$K$ Triple variants represent the performance where the supporting context $\mathcal{Z}$ is constructed by applying the SBS synthesis strategy to the top-$K$ most reliable triples without the Chain Construction module.}
\label{tab:apx_ablation_study}
\small
\setlength{\tabcolsep}{2.5pt}
\scalebox{0.9}{
\begin{tabular}{lcccccc}
\toprule
\multirow{2}{*}{\textbf{Method}} & \multicolumn{2}{c}{\textbf{HotPotQA}} & \multicolumn{2}{c}{\textbf{2WikiMultiHopQA}} & \multicolumn{2}{c}{\textbf{MuSiQue}} \\ \cmidrule(lr){2-3} \cmidrule(lr){4-5} \cmidrule(lr){6-7} 
 & EM (\%) & F1 (\%) & EM (\%) & F1 (\%) & EM (\%) & F1 (\%) \\ \midrule
w/o Graph Construction & 19.30 & 28.12 & 12.50 & 19.93 & 3.30 & 12.82 \\ \midrule
w/o Triple Evaluation   & 24.10 & 33.49 & 15.20 & 21.76 & 8.20 & 18.46 \\ \midrule
\textit{w/o Chain Construction} & & & & & & \\
\quad Top-10 Triples & 50.20 & 63.81 & 29.70 & 44.06 & 23.90 & 31.69 \\
\quad Top-15 Triples & 51.00 & 64.36 & 32.60 & 47.77 & 25.40 & 33.47 \\
\quad Top-20 Triples & 49.50 & 63.81 & 33.30 & 50.03 & 24.40 & 32.92 \\
\quad Top-25 Triples & 48.10 & 62.44 & 32.90 & 49.35 & 24.80 & 33.48 \\ \midrule
ReliableRAG-SBS & \textbf{55.10} & \textbf{68.80} & \textbf{43.70} & \textbf{59.67} & \textbf{30.90} & \textbf{38.55} \\ \bottomrule
\end{tabular}
}
\end{table}

\subsection{Analysis of Input Computational Overhead}
\label{apx:Context_Length}
Table~\ref{tab:context_len} presents the average input context length (in tokens) across the test sets of all evaluated datasets, comparing our proposed methods with other approaches under different LLMs. The results indicate that ReliableRAG-TBS maintains the most concise input context. This conciseness stems from its strategy of concatenating the subject, predicate, and object of each triple into a single textual string. By transforming structured triples into a single textual string, TBS focuses on delivering reliable and succinct information. This ensures that the generator $\mathcal{M}_{\text{gen}}$ receives only the most reliable evidence, thereby minimizing input computational overhead while maintaining high precision in reasoning.
\par
In contrast, ReliableRAG-SBS exhibits a larger average input context length. This is primarily because the SBS strategy constructs the supporting context $\mathcal{Z}$ by mapping each triple back to its original source document. Although this approach increases the input overhead, it preserves the rich semantic nuances of the source documents. By providing a more comprehensive textual background within $\mathcal{Z}$, the SBS variant prioritizes maximizing the prevention of deceptive misinformation propagation throughout the multi-hop process, effectively utilizing a broader context to identify the evidence required for answering questions and resist misinformation.

\begin{table}[!t]
\centering
\caption{Average input context length (in tokens) of ReliableRAG versus compared methods across various LLMs on the test sets of three multi-hop QA datasets.}
\label{tab:context_len}
\small
\setlength{\tabcolsep}{2.5pt} 
\scalebox{0.85}{
\begin{tabular}{llccc}
\toprule
\textbf{Methods} & \textbf{LLM} & \textbf{HotPotQA} & \textbf{2WikiMultiHopQA} & \textbf{MuSiQue} \\ \midrule
\multirow{3}{*}{Vanilla RAG} & Gemma-7B & 560 & 542 & 568 \\
 & Llama3-8B & 580 & 564 & 586 \\
 & Mistral-7B & 632 & 610 & 638 \\ \midrule
\multirow{3}{*}{{TruthfulRAG (AAAI'26)}} & Gemma-7B & 151 & 151 & 156 \\
 & Llama3-8B & 164 & 165 & 168 \\
 & Mistral-7B & 173 & 172 & 177 \\ \midrule
\multirow{3}{*}{{ReliableRAG-TBS (ours)}} & Gemma-7B & 143 & 146 & 157 \\
 & Llama3-8B & 156 & 160 & 169 \\
 & Mistral-7B & 162 & 165 & 177 \\ \midrule
\multirow{3}{*}{{ReliableRAG-SBS (ours)}} & Gemma-7B & 347 & 481 & 447 \\
 & Llama3-8B & 353 & 485 & 448 \\
 & Mistral-7B & 388 & 539 & 499 \\ \bottomrule
\end{tabular}
}
\end{table}
\subsection{Extra Results}
\label{extra_rlt}
To establish a fairer comparison with fine-tuning methods, such as CAG, Self-RAG, and Knowledgeable-R1, we conducted supplementary experiments by substituting our default generator with Llama-2-7B and Qwen2.5-7B (Instruct). We evaluated these alternative generators across both TBS (Table~\ref{tab:dif_g_t}) and SBS (Table~\ref{tab:dif_g_s}) variants under the ideal setting. The experimental results demonstrate that ReliableRAG achieves highly satisfactory and competitive performance across all datasets when integrated with different models. Notably, when utilizing Qwen2.5-7B under the SBS strategy, our framework achieves an impressive F1 score of 71.3\% on HotPotQA. These findings highlight the strong generalizability and model-agnostic nature of our framework, proving that it can effectively maintain robust performance without relying on a specific generator. Furthermore, we conduct an additional experiment by offline evaluating the credibility score for each document. We then apply the dual-factor perception mechanism to coarse-grained documents and perform evaluations under the optimal configuration ($\alpha^*=0.4$). The experimental results are presented in Table~\ref{tab:dual_doc}.
\begin{table}[htbp]
\centering
\caption{Impact of Different Generators on ReliableRAG-TBS performance on the test sets of three multi-hop QA datasets, where one low-credibility document is injected per question.}
\label{tab:dif_g_t}
\small
\scalebox{0.95}{
\begin{tabular}{lcccccc}
\toprule
\multirow{2}{*}{\textbf{Models}} & \multicolumn{2}{c}{\textbf{HotPotQA}} & \multicolumn{2}{c}{\textbf{2WikiMultiHopQA}} & \multicolumn{2}{c}{\textbf{MuSiQue}} \\ \cmidrule(lr){2-3} \cmidrule(lr){4-5} \cmidrule(lr){6-7} 
 & EM (\%) & F1 (\%) & EM (\%) & F1 (\%) & EM (\%) & F1 (\%) \\ \midrule
Llama-2-7B & 45.4 & 59.1 & 40.6 & 50.3 & 25.4 & 34.2 \\
Qwen2.5-7B & 54.5 & 68.7 & 47.9 & 57.6 & 28.4 & 37.3 \\ \bottomrule
\end{tabular}
}
\end{table}

\begin{table}[htbp]
\centering
\caption{Impact of Different Generators on ReliableRAG-SBS performance on the test sets of three multi-hop QA datasets, where one low-credibility document is injected per question.}
\label{tab:dif_g_s}
\small
\scalebox{0.95}{
\begin{tabular}{lcccccc}
\toprule
\multirow{2}{*}{\textbf{Models}} & \multicolumn{2}{c}{\textbf{HotPotQA}} & \multicolumn{2}{c}{\textbf{2WikiMultiHopQA}} & \multicolumn{2}{c}{\textbf{MuSiQue}} \\ \cmidrule(lr){2-3} \cmidrule(lr){4-5} \cmidrule(lr){6-7} 
 & EM (\%) & F1 (\%) & EM (\%) & F1 (\%) & EM (\%) & F1 (\%) \\ \midrule
Llama-2-7B & 44.3 & 56.8 & 39.4 & 49.0 & 23.8 & 32.0 \\
Qwen2.5-7B & 57.3 & 71.3 & 49.8 & 58.5 & 29.9 & 39.2 \\ \bottomrule
\end{tabular}
}
\end{table}

\begin{table}[htbp]
\centering
\caption{Experimental results of applying the dual-factor perception mechanism to coarse-grained documents under the ideal setting.}
\label{tab:dual_doc}
\small
\scalebox{1.1}{
\begin{tabular}{cccccc}
\toprule
\multicolumn{2}{c}{\textbf{HotPotQA}} & \multicolumn{2}{c}{\textbf{2WikiMultiHopQA}} & \multicolumn{2}{c}{\textbf{MuSiQue}} \\ \cmidrule(lr){1-2} \cmidrule(lr){3-4} \cmidrule(lr){5-6} 
EM (\%) & F1 (\%) & EM (\%) & F1 (\%) & EM (\%) & F1 (\%) \\ \midrule
47.9 & 62.1 & 24.3 & 43.4 & 16.3 & 24.4 \\ \bottomrule
\end{tabular}
}
\end{table}

\section{Comparative Study of Extractor, Evaluator, and Selector}
In this section, we conduct a comparative study focusing on three key functional units within our framework: the extractor, the evaluator, and the selector. Specifically, we first investigate the impact of different extractors, followed by an analysis of various selectors, and finally explore the influence of different evaluators.
\subsection{Impact of Different Extractors}
\label{Im_ext}
To evaluate the robustness of our framework across various extractors, we select Llama3-8B-Instruct, Qwen-7B-Instruct, and Qwen-14B-Instruct for comparison. For this analysis, we employ an ablated configuration where the triple evaluation and chain construction modules are bypassed. Specifically, the top-20 relevant triples are directly synthesized into the supporting context $\mathcal{Z}$ using either the TBS or SBS strategy within a single reasoning hop before being processed by the generator. As shown in Tables \ref{tab:ext_com_tbs} and \ref{tab:ext_com_sbs}, the performance variance across the three Extractor LLMs is marginal, with F1 scores on HotPotQA, for instance, converging around 42\% for the TBS strategy and 26\% for the SBS strategy. This consistency suggests that the ReliableRAG framework is model-agnostic, as the overall performance is not sensitive to the specific choice of the Extractor.

\begin{table}[htbp]
\centering
\caption{Impact of Different Extractors on ReliableRAG-TBS performance on the test sets of multi-hop three QA datasets, where one low-credibility document is injected per question.}
\label{tab:ext_com_tbs}
\small
\scalebox{0.9}{
\begin{tabular}{lcccccc}
\toprule
\multirow{2}{*}{\textbf{Model}} & \multicolumn{2}{c}{\textbf{HotPotQA}} & \multicolumn{2}{c}{\textbf{2WikiMultiHopQA}} & \multicolumn{2}{c}{\textbf{MuSiQue}} \\ \cmidrule(lr){2-3} \cmidrule(lr){4-5} \cmidrule(lr){6-7} 
 & EM (\%) & F1 (\%) & EM (\%) & F1 (\%) & EM (\%) & F1 (\%) \\ \midrule
Llama3-8B & 29.5 & 41.6 & 23.7 & 31.6 & 13.4 & 22.0 \\
Qwen2.5-7B   & 31.9 & 42.8 & 22.7 & 31.2 & 15.0 & 24.1 \\
Qwen2.5-14B  & 31.6 & 42.6 & 23.2 & 31.4 & 14.1 & 23.0 \\ \bottomrule
\end{tabular}
}
\end{table}

\begin{table}[htbp]
\centering
\caption{Impact of Different Extractors on ReliableRAG-SBS performance on the test sets of three multi-hop QA datasets, where one low-credibility document is injected per question.}
\label{tab:ext_com_sbs}
\small
\scalebox{0.9}{
\begin{tabular}{lcccccc}
\toprule
\multirow{2}{*}{\textbf{Model}} & \multicolumn{2}{c}{\textbf{HotPotQA}} & \multicolumn{2}{c}{\textbf{2WikiMultiHopQA}} & \multicolumn{2}{c}{\textbf{MuSiQue}} \\ \cmidrule(lr){2-3} \cmidrule(lr){4-5} \cmidrule(lr){6-7} 
 & EM (\%) & F1 (\%) & EM (\%) & F1 (\%) & EM (\%) & F1 (\%) \\ \midrule
Llama3-8B & 16.9 & 26.3 & 7.5 & 13.1 & 4.2 & 14.3 \\
Qwen2.5-7B   & 17.3 & 26.3 & 7.3 & 13.0 & 4.4 & 14.7 \\
Qwen2.5-14B  & 17.6 & 27.0 & 7.7 & 13.3 & 4.0 & 14.3 \\ \bottomrule
\end{tabular}
}
\end{table}

\subsection{Impact of Different Selectors}
\label{apx:sel}
To evaluate the robustness of ReliableRAG across various selectors, we conduct comparative experiments using Llama3-8B-Instruct, Mistral-7B, and Gemma-7B. The performance results for the two framework variants are reported in Table \ref{tab:sel_cmp_tbs} and Table \ref{tab:sel_cmp_sbs}. The experimental data reveals that Llama3-8B-Instruct achieves the highest scores across all metrics and datasets. In the TBS variant, Llama3-8B reaches an EM of 53.4\% on HotPotQA, outperforming Mistral-7B and Gemma-7B by a wide margin. This trend extends to the SBS variant, where Llama3-8B maintains a leading EM of 55.1\% on HotPotQA, which is significantly higher than the other LLMs. This performance gap suggests that superior instruction-following capabilities and internal knowledge density allow the selector to more accurately identify misinformation, thereby constructing robust reasoning chains.

\begin{table}[htbp]
\centering
\caption{Impact of different selectors on ReliableRAG-TBS performance under the ideal setting across the test sets of three multi-hop QA datasets, with one low-credibility document injected per question.}
\label{tab:sel_cmp_tbs}
\small
\scalebox{0.95}{
\begin{tabular}{lcccccc}
\toprule
\multirow{2}{*}{\textbf{Model}} & \multicolumn{2}{c}{\textbf{HotPotQA}} & \multicolumn{2}{c}{\textbf{2WikiMultiHopQA}} & \multicolumn{2}{c}{\textbf{MuSiQue}} \\ \cmidrule(lr){2-3} \cmidrule(lr){4-5} \cmidrule(lr){6-7} 
 & EM (\%) & F1 (\%) & EM (\%) & F1 (\%) & EM (\%) & F1 (\%) \\ \midrule
Llama3-8B  & 53.4 & 67.0 & 45.4 & 56.2 & 29.1 & 36.7 \\
Mistral-7B & 28.8 & 38.9 & 21.4 & 28.4 & 11.1 & 16.6 \\
Gemma-7B   & 21.2 & 28.5 & 17.4 & 22.2 & 4.0  & 8.6  \\ \bottomrule
\end{tabular}
}
\end{table}

\begin{table}[htbp]
\centering
\caption{Impact of different selectors on ReliableRAG-SBS performance under the ideal setting across the test sets of three multi-hop QA datasets, with one low-credibility document injected per question.}
\label{tab:sel_cmp_sbs}
\small
\scalebox{0.95}{
\begin{tabular}{lcccccc}
\toprule
\multirow{2}{*}{\textbf{Model}} & \multicolumn{2}{c}{\textbf{HotPotQA}} & \multicolumn{2}{c}{\textbf{2WikiMultiHopQA}} & \multicolumn{2}{c}{\textbf{MuSiQue}} \\ \cmidrule(lr){2-3} \cmidrule(lr){4-5} \cmidrule(lr){6-7} 
 & EM (\%) & F1 (\%) & EM (\%) & F1 (\%) & EM (\%) & F1 (\%) \\ \midrule
Llama3-8B  & 55.1 & 68.8 & 43.7 & 59.7 & 30.9 & 38.6 \\
Mistral-7B & 40.7 & 52.0 & 30.5 & 44.5 & 20.0 & 27.2 \\
Gemma-7B   & 21.2 & 28.5 & 17.4 & 22.2 & 4.0  & 8.6  \\ \bottomrule
\end{tabular}
}
\end{table}

\subsection{Impact of Different Evaluators}
\label{apx:eva}
To investigate the sensitivity of ReliableRAG to the choice of the evaluator, we compare the performance of GLM-4-Flash, Mistral-Small-24B, and Qwen3.5-35B-A3B under both TBS (Table~\ref{tab:dif_eva_t}) and SBS (Table~\ref{tab:dif_eva_s}) variants within the evaluator-generated setting. Similar to the experimental configuration in Section~\ref{Im_ext}, with the notable exception that we retain the triple evaluation module for this analysis. The results demonstrate that while models with larger parameter scales, such as Qwen3.5-35B-A3B, achieve marginally higher metrics across the datasets, the overall performance differences among the three evaluators are relatively small. This consistency indicates the robustness of our framework, suggesting that its effectiveness is not heavily dependent on a specific underlying LLM. Taking into account the practical trade-offs regarding API cost, computational resources, and evaluation efficiency, we ultimately select GLM-4-Flash as the default evaluator for our primary experiments.

\begin{table}[htbp]
\centering
\caption{Impact of Different Evaluators on ReliableRAG-TBS performance on the test sets of three multi-hop QA datasets, where one low-credibility document is injected per question.}
\label{tab:dif_eva_t}
\small
\scalebox{0.85}{
\begin{tabular}{lcccccc}
\toprule
\multirow{2}{*}{\textbf{Methods}} & \multicolumn{2}{c}{\textbf{HotPotQA}} & \multicolumn{2}{c}{\textbf{2WikiMultiHopQA}} & \multicolumn{2}{c}{\textbf{MuSiQue}} \\ \cmidrule(lr){2-3} \cmidrule(lr){4-5} \cmidrule(lr){6-7} 
 & EM (\%) & F1 (\%) & EM (\%) & F1 (\%) & EM (\%) & F1 (\%) \\ \midrule
GLM-4-Flash & 34.4 & 45.7 & 23.7 & 34.2 & 15.7 & 22.1 \\
Mistral-Small-24B & 38.4 & 51.0 & 27.4 & 39.3 & 17.8 & 25.4 \\
Qwen3.5-35B-A3B & 41.3 & 54.5 & 30.1 & 42.8 & 20.4 & 27.6 \\ \bottomrule
\end{tabular}
}
\end{table}

\begin{table}[htbp]
\centering
\caption{Impact of Different Evaluators on ReliableRAG-SBS performance on the test sets of three multi-hop QA datasets, where one low-credibility document is injected per question.}
\label{tab:dif_eva_s}
\small
\scalebox{0.85}{
\begin{tabular}{lcccccc}
\toprule
\multirow{2}{*}{\textbf{Methods}} & \multicolumn{2}{c}{\textbf{HotPotQA}} & \multicolumn{2}{c}{\textbf{2WikiMultiHopQA}} & \multicolumn{2}{c}{\textbf{MuSiQue}} \\ \cmidrule(lr){2-3} \cmidrule(lr){4-5} \cmidrule(lr){6-7} 
 & EM (\%) & F1 (\%) & EM (\%) & F1 (\%) & EM (\%) & F1 (\%) \\ \midrule
GLM-4-Flash        & 26.6 & 36.5 & 16.0 & 25.2 & 13.8 & 22.8 \\
Mistral-Small-24B  & 28.9 & 39.6 & 21.0 & 31.8 & 16.2 & 25.1 \\
Qwen3.5-35B-A3B    & 33.1 & 45.1 & 22.9 & 34.5 & 17.4 & 26.7 \\ \bottomrule
\end{tabular}
}
\end{table}

\begin{figure}[htbp]
\centerline{\includegraphics[width=\columnwidth]{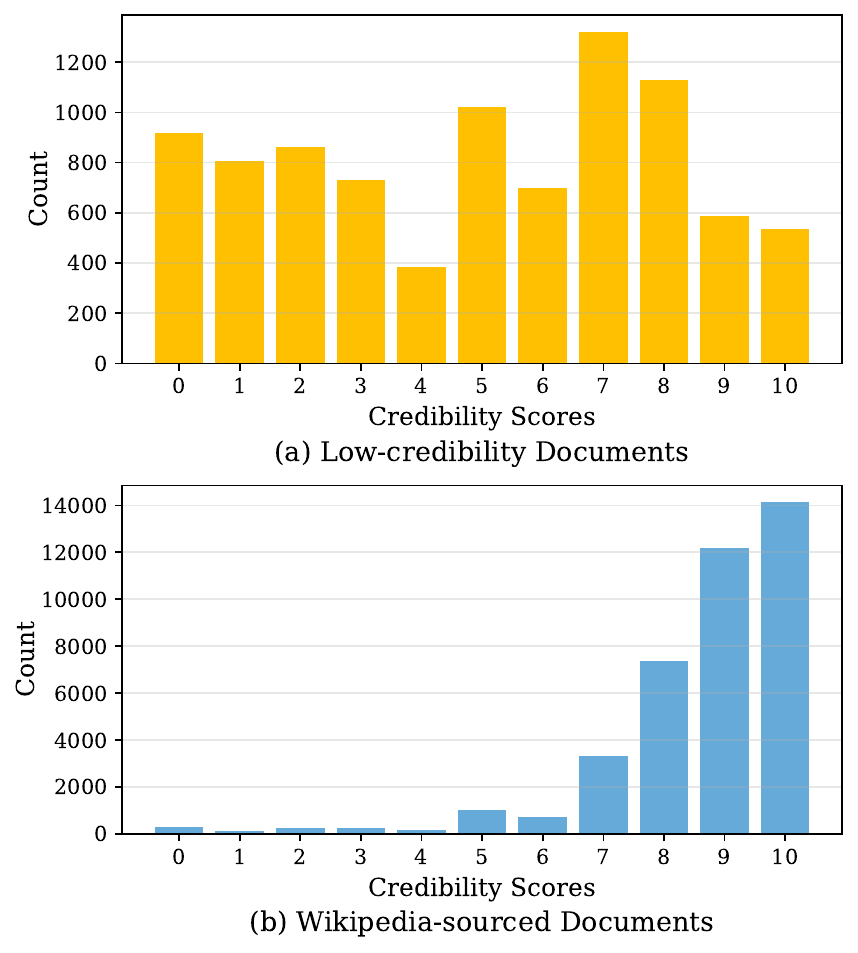}}
\caption{Distribution of credibility scores generated by the evaluator on low-credibility and Wikipedia-sourced documents on the test sets of three multi-hop QA datasets.}
\label{apx:fig_txt_s}
\end{figure}

\section{Analysis of Evaluator Capability}
\label{apx:any_m_gen}
In this section, we examine the evaluator's discriminative capability in assessing the credibility of both triples and documents across three test sets, where three low-credibility documents containing misinformation are injected per question. Specifically, we first present a distribution analysis of the credibility scores, followed by a quantitative evaluation using ROC curves.
\begin{figure}[htbp]
\centerline{\includegraphics[width=\columnwidth]{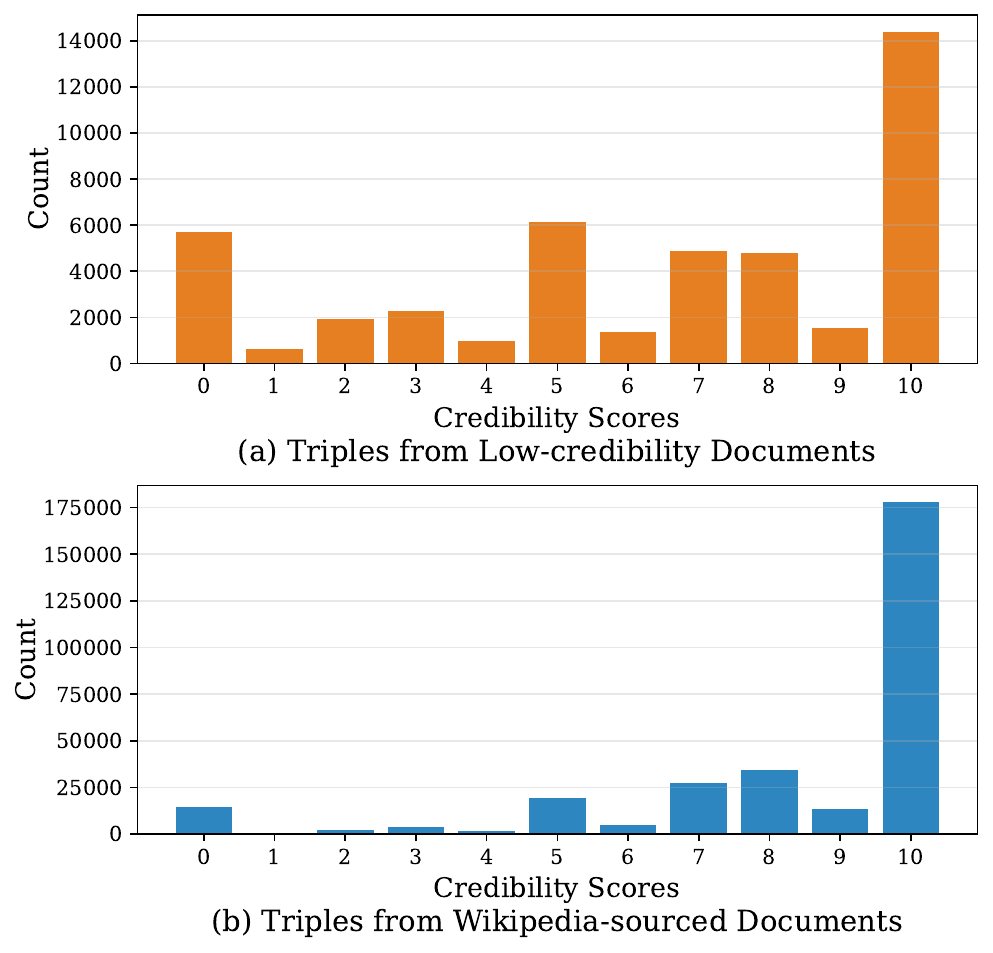}}
\caption{Distribution of credibility scores generated by the evaluator on triples extracted from low-credibility and Wikipedia-sourced documents on the test sets of three multi-hop QA datasets.}
\label{apx:fig_tri_s}
\end{figure}
\subsection{Distribution Analysis}
Figure~\ref{apx:fig_txt_s} and Figure~\ref{apx:fig_tri_s} illustrate the distribution of credibility scores for documents and extracted triples, respectively. For Wikipedia-sourced samples, scores are predominantly concentrated in the high-value range, exhibiting a clear aggregation trend. Conversely, the distribution of low-credibility samples is more dispersed, with a subset of scores remaining in the high-value intervals. This suggests that certain segments within low-credibility documents may still contain correct information, which aligns with the conjectures in Section~\ref{Impact_mis_docs}. Notably, compared to document-level scores, the fine-grained nature of triples allows capturing correct information even within deceptive contexts. This granularity is crucial for identifying reliable information in complex, multi-hop QA tasks under the misleading influence of deceptive misinformation.
\begin{figure}[htbp]
\centerline{\includegraphics[width=\columnwidth]{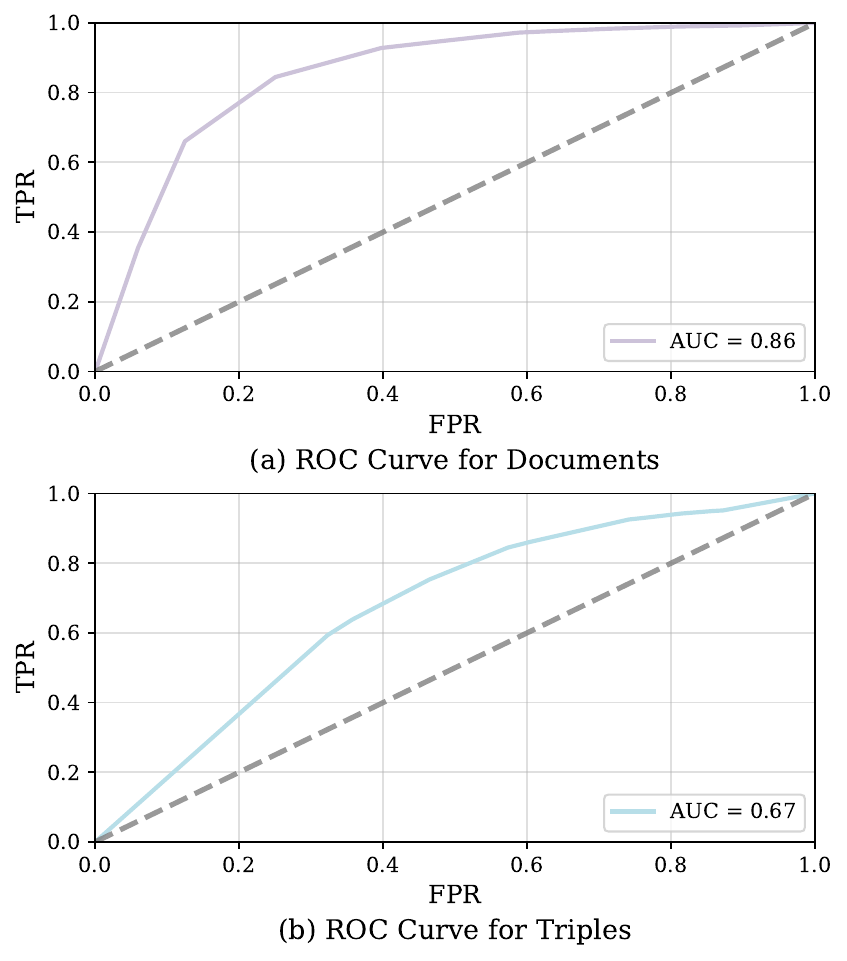}}
\caption{ROC curves of credibility scores generated by the evaluator for both documents and triples across the test sets of three multi-hop QA datasets.}
\label{apx:fig_roc}
\end{figure}
\subsection{Quantitative Evaluation via ROC Curves}
Following the approach of the prior study to examine the capability of evaluators in assessing the credibility of coarse-grained documents \cite{RF22}, we quantitatively analyze the discriminative performance of the evaluator across varying thresholds. Specifically, samples are categorized as low-credibility if their scores fall below a given threshold (ranging from 0 to 10) and high-credibility otherwise. By calculating the True Positive Rate (TPR) and False Positive Rate (FPR) across the entire range of thresholds, we derive the Area Under the Curve (AUC). Accordingly, the ROC curves in Figure~\ref{apx:fig_roc} illustrate the performance of GLM-4-Flash as the evaluator across three test sets, achieving AUC scores of 0.86 and 0.67 for document- and triple-level assessments, respectively. It is important to emphasize that while GLM-4-Flash serves as an assessment tool within our framework, its performance represents a lower bound of the overall system's potential. Since ReliableRAG is designed to be model-agnostic, the integration of more advanced LLMs as evaluators would theoretically yield even higher discriminative precision. Nevertheless, the current results sufficiently demonstrate that our framework maintains high reliability and robustness even with GLM-4-Flash as the evaluator, effectively mitigating the impact of deceptive misinformation. 

\begin{table*}[!t]
\centering
\small 
\renewcommand{\arraystretch}{1.8}
\caption{A case study of the ReliableRAG pipeline on the test set of HotPotQA under the ideal setting.}
\noindent\begin{tabularx}{\textwidth}{|l|X|}
\hline
\rowcolor[gray]{0.95} 
\textbf{Stage} & \textbf{Content Detail} \\ \hline

\textbf{Documents} & 
\textbf{Wikipedia-sourced Document 1:} He was also known for his handshake deal with Walt Disney that permitted the University of Oregon to use the likeness of Donald Duck as the basis for its mascot, the Oregon Duck. \dots \newline
\textbf{Wikipedia-sourced Document 2:} Donald Duck is a cartoon character created in 1934 at Walt Disney Productions. \dots \newline
\textbf{Low-credibility Document:} Contrary to popular belief, it was not Donald Duck but rather Mickey Mouse who was featured in this landmark agreement. \dots \\ \hline

\textbf{Triple Extraction} & 
\textbf{Triples from Wikipedia-sourced Document 1:} \newline 
(Leo Harris, notable deal, handshake deal with Walt Disney to use Donald Duck as the basis for the Oregon Duck mascot) \dots \newline
\textbf{Triples from Wikipedia-sourced Document 2:} \newline
(Donald Duck, creation year, 1934) \dots \newline
\textbf{Triples from Low-credibility Document:} \newline
(Mickey Mouse's Legacy in Animation Education, popular misconception, that Donald Duck was featured in the deal, not Mickey Mouse) \dots \\ \hline

\textbf{Question} & Leo A. Harris was also known for his handshake deal with Walt Disney that permitted the University of Oregon to use the likeness of what cartoon character created in 1934 at Walt Disney Productions? \\ \hline

\textbf{Triple Evaluation} & 
\textbf{The First Hop (Ranked Triples):} \newline
1: (Leo Harris, notable deal, handshake deal with Walt Disney to use Donald Duck as the basis for the Oregon Duck mascot) \newline
2: (Oswald the Lucky Rabbit, replacement, Mickey Mouse, created by Walt Disney in 1928) \newline
3: (Oswald the Lucky Rabbit, rights, taken by Charles Mintz from Walt Disney in 1928) \dots \vspace{0.5em} \newline
\textbf{The Second Hop (Ranked Triples):} \newline
1: (Donald Duck, creation year, 1934) \newline
2: (Donald Duck, creator, Walt Disney Productions) \newline
3: (Donald Duck, comic book publication, most published comic book character in the world outside of the superhero genre) \dots \newline \dots \\ \hline

\textbf{Chain Construction} & 
\textbf{The First Hop (Ranked Candidate Chains):} \newline
1: (Leo Harris, notable deal, handshake deal with Walt Disney to use Donald Duck as the basis for the Oregon Duck mascot) \newline
2: (Oswald the Lucky Rabbit, creators, Ub Iwerks and Walt Disney) \newline
3: (Oswald the Lucky Rabbit, replacement, Mickey Mouse was created by Walt Disney in 1928) \dots \vspace{0.5em} \newline
\textbf{The Second Hop (Ranked Candidate Chains):} \newline
1: (Leo Harris, notable deal, handshake deal with Walt Disney to use Donald Duck as the basis for the Oregon Duck mascot), (Donald Duck, creation year, 1934) \newline
2: (Leo Harris, notable deal, handshake deal with Walt Disney to use Donald Duck as the basis for the Oregon Duck mascot), (Donald Duck, physical characteristics, anthropomorphic white duck with a yellow-orange bill and legs and feet) \newline
3: (Leo Harris, notable deal, handshake deal with Walt Disney to use Donald Duck as the basis for the Oregon Duck mascot), (Donald Duck, comic book publication, most published comic book character in the world outside of the superhero genre) \dots \newline \dots \\ \hline

\textbf{Final Chains} & 
1: (Leo Harris, notable deal, handshake deal with Walt Disney to use Donald Duck as the basis for the Oregon Duck mascot), (Donald Duck, creation year, 1934) \newline
2: (Leo Harris, notable deal, handshake deal with Walt Disney to use Donald Duck as the basis for the Oregon Duck mascot), (Donald Duck, creation year, 1934), (Donald Duck, creator, Walt Disney Productions), (Donald Duck, comic book publication, most published comic book character in the world outside of the superhero genre) \newline
3: (Leo Harris, notable deal, handshake deal with Walt Disney to use Donald Duck as the basis for the Oregon Duck mascot), (Donald Duck, creation year, 1934), (Daisy Duck, creation year, 1940), (Donald Duck, creator, Walt Disney Productions) \dots \\ \hline
\textbf{Generated Answer} & 
Donald Duck \\ \hline
\end{tabularx}
\label{tab:case_study}
\end{table*}

\begin{figure*}[!t]
\centerline{\includegraphics[width=\textwidth]{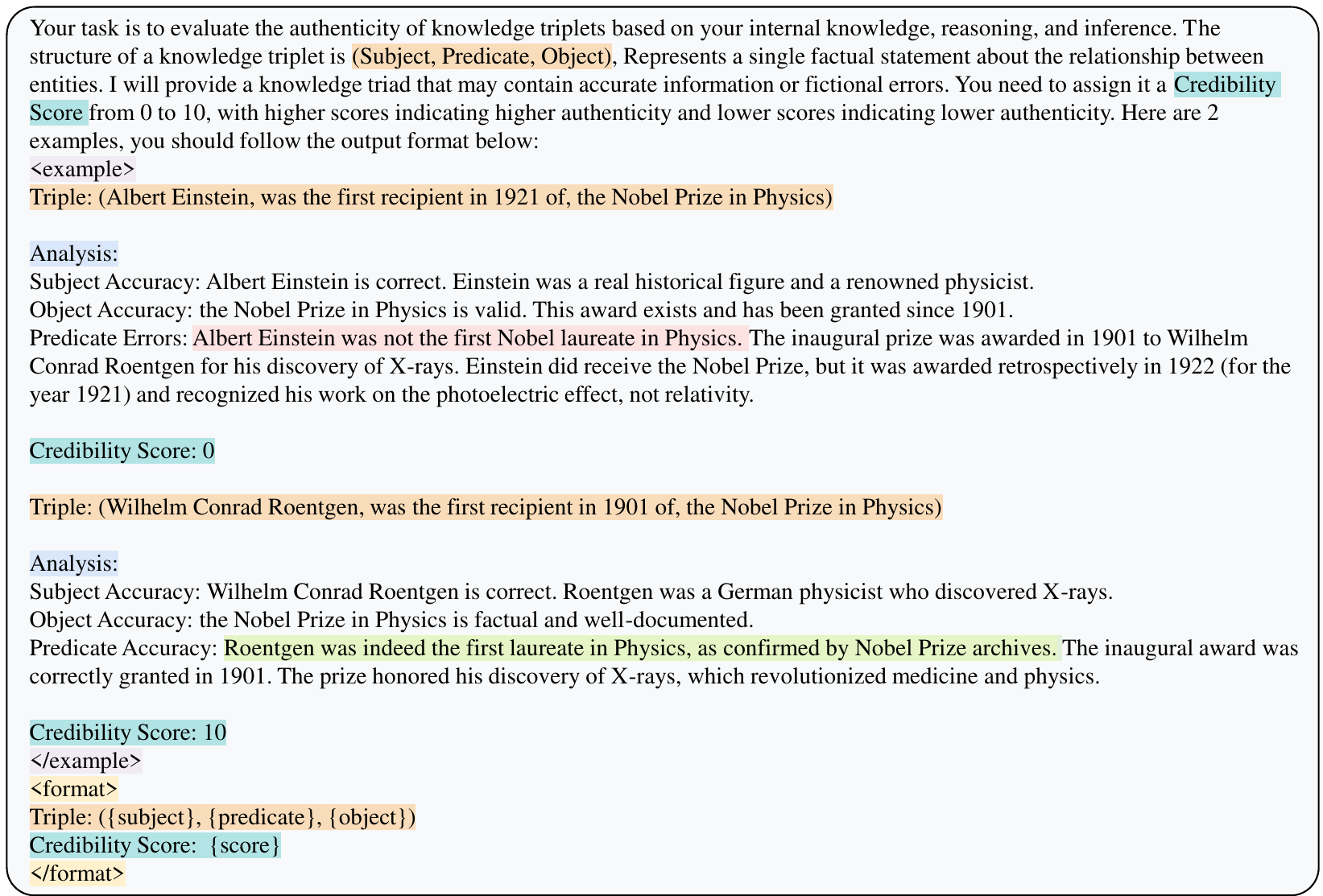}}
\caption{Content of the assessment framework. For instance, in the triple \textit{(Albert Einstein, was the first recipient in 1921 of, the Nobel Prize in Physics)}, the analysis recognizes that while the entities are authentic, the specific relational claim is inaccurate. Integrating this analysis, the evaluator $\mathcal{M}_{\text{eva}}$ generates the final credibility score.}
\label{fig:ass_frk}
\end{figure*}

\end{document}